\documentclass{article}

    \PassOptionsToPackage{numbers, compress}{natbib}

\usepackage[final]{neurips_2024}

\usepackage[utf8]{inputenc} % allow utf-8 input
\usepackage[T1]{fontenc}    % use 8-bit T1 fonts
\usepackage{hyperref}       % hyperlinks
\usepackage{url}            % simple URL typesetting
\usepackage{booktabs}       % professional-quality tables
\usepackage{amsfonts}       % blackboard math symbols
\usepackage{nicefrac}       % compact symbols for 1/2, etc.
\usepackage{microtype}      % microtypography
\usepackage{xcolor}         % colors
\usepackage{graphicx}
\usepackage{multicol}
\usepackage{multirow}
\usepackage{amsmath}
\usepackage{amsthm}
\usepackage{amssymb}
\usepackage{tcolorbox}
\usepackage{enumitem}
\usepackage{wrapfig}
\usepackage{caption}

\newtcbox{\mybox}[1][red]
  {on line, arc = 0pt, outer arc = 0pt,
    colback = #1!10!white, colframe = #1!50!black,
    boxsep = 0pt, left = 1pt, right = 1pt, top = 1pt, bottom = 1pt,
    boxrule = 0pt, bottomrule = 1pt, toprule = 1pt, fontupper = \ttfamily}

\title{Meaningful Learning: Enhancing Abstract Reasoning in Large Language Models via Generic Fact Guidance}

\author{
Kai Xiong$^\spadesuit$\;\; Xiao Ding$^\spadesuit$\thanks{Corresponding Authors}\;\;\; Ting Liu$^\spadesuit$\;\; Bing Qin$^\spadesuit$\\ \textbf{Dongliang Xu}$^\blacklozenge$\;\; \textbf{Qing Yang}$^\blacklozenge$\;\; \textbf{Hongtao Liu}$^\blacklozenge$\;\; \textbf{Yixin Cao}$^{\clubsuit*}$ \\ $^\spadesuit$\footnotesize Research Center for Social Computing and Information Retrieval\\\footnotesize Harbin Institute of Technology, Harbin, China \\
$^\clubsuit$\footnotesize Fudan University, Shanghai, China\\$^\blacklozenge$\footnotesize Du Xiaoman (Beijing) Science Technology Co., Ltd., Beijing, China \\
\footnotesize\texttt{\{kxiong, xding, tliu, qinb\}@ir.hit.edu.cn}\\
\footnotesize\texttt{\{xudongliang, yangqing, liuhongtao01\}@duxiaoman.com}\\
\footnotesize\texttt{yxcao@fudan.edu.cn}
}

\begin{document}

\maketitle

\begin{abstract}
Large language models~(LLMs) have developed impressive performance and strong explainability across various reasoning scenarios, marking a significant stride towards mimicking human-like intelligence. Despite this, when tasked with several simple questions supported by a generic fact, LLMs often struggle to abstract and apply the generic fact to provide consistent and precise answers, revealing a deficiency in abstract reasoning abilities. This has sparked a vigorous debate about whether LLMs are genuinely reasoning or merely memorizing. In light of this, we design a preliminary study to quantify and delve into the abstract reasoning abilities of existing LLMs. Our findings reveal a substantial discrepancy between their general reasoning and abstract reasoning performances. To relieve this problem, we tailor an abstract reasoning dataset~(AbsR) together with a meaningful learning paradigm to teach LLMs how to leverage generic facts for reasoning purposes. The results show that our approach not only boosts the general reasoning performance of LLMs but also makes considerable strides towards their capacity for abstract reasoning, moving beyond simple memorization or imitation to a more nuanced understanding and application of generic facts. The code is available at \href{https://github.com/Waste-Wood/MeanLearn}{\texttt{https://github.com/Waste-Wood/MeanLearn}}.
\end{abstract}

\section{Introduction}
% Humans proceed meaningful learning to induce common patterns or high-level abstractions to aquire abstract reasoning abilities~\cite{lake2017building,qiu2023phenomenal}. Such capabilities enable humans to apply a generic fact~(e.g. \emph{copper is a good therminal conductor}) across a variety of problems, exemplifying a profound level of understanding and application of knowledge. As shown in Figure~\ref{fig:intro} (a), humans can deduce ``rock dissolved'' and ``solution turns orange'' when given ``adding rock into hydrochloric acid'' and ``adding rust into sulphuric acid'', respectively. This stems from the established expertise in human mind~(the generic fact ``acid is corrosive'' behind the two questions), and extends to applications in different scenarios.

\begin{figure}[t]
\centering
\begin{minipage}[l]{0.5\textwidth}
    \centering
    \includegraphics[width=1.0\columnwidth]{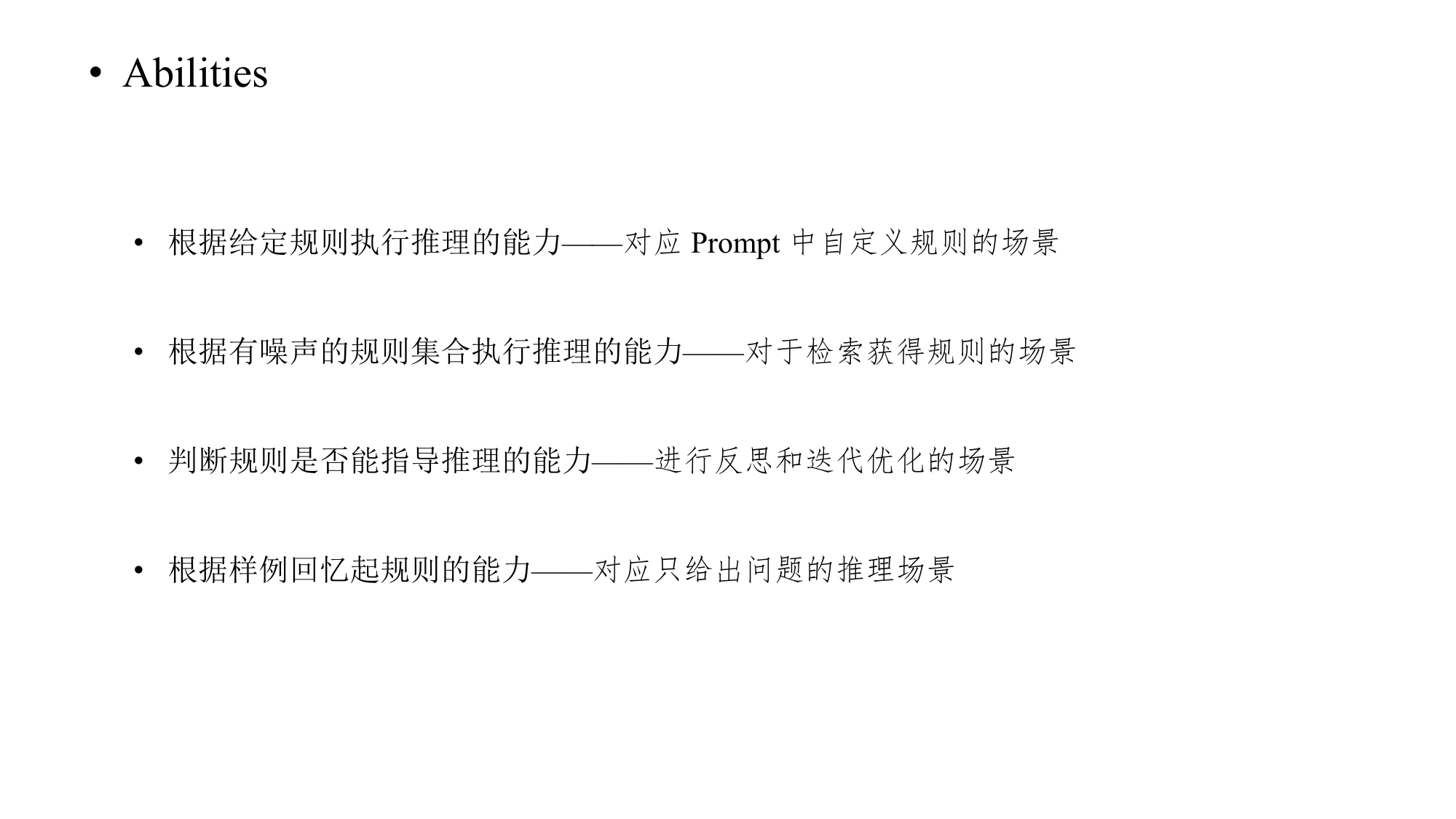}
    \caption{The responses of (a) Humans and (b) LLMs when facing two questions which are supported by the same generic fact.}
    \label{fig:intro}
\end{minipage}
\quad
\begin{minipage}[r]{0.46\textwidth}
    \small
    \centering
    \setlength\tabcolsep{2.2pt}
    \captionof{table}{The vanilla accuracy and abstract reasoning accuracy~(\texttt{AbsAcc}) on e-CARE~\cite{du2022care}. We will formally define \texttt{AbsAcc} in Sec.~\ref{sec:metric}.}
    \begin{tabular}{lccc}
    \toprule
    \textbf{Methods} & Size & Vanilla Accuracy & \texttt{AbsAcc} \\ \midrule
    \multirow{2}{*}{LLaMA-2~\cite{touvron2023llama}} & 7B   & 51.67   & 24.82    \\
                         & 13B  & 68.45   & 42.94    \\ \midrule
    \multirow{2}{*}{Orca-2~\cite{mitra2023orca}}  & 7B   & 74.23   & 51.19    \\
                         & 13B  & 79.83   & 59.47    \\ \midrule
    GPT-3.5~\cite{brown2020language} & >20B & 83.25 & 66.08 \\ \midrule
    Human~\cite{du2022care} & -- & 92.00 & 89.90 \\\bottomrule
    \end{tabular}
    \label{tab:intro}
\end{minipage}
% \vspace{-0.35cm}
\end{figure}

Humans proceed with meaningful learning to induce common patterns or high-level abstractions to acquire abstract reasoning abilities~\cite{lake2017building,qiu2023phenomenal}. Such capabilities allow us to apply broad principles across diverse situations, demonstrating a deep understanding and versatile application of knowledge. As shown in Figure~\ref{fig:intro}~(a), humans can deduce ``rock dissolved'' and ``the skin suffers pain'' when given ``adding rock into hydrochloric acid'' and ``acid touches human skin'', respectively. This stems from the established expertise in the human mind~(the generic fact ``acid is corrosive''), and extends to applications in different scenarios. Of late, remarkable headways of LLMs have pushed AI much further towards human-like intelligence~\cite{achiam2023gpt,touvron2023llama}. Interestingly, can LLMs act like humans to instinctively consider the generic fact for flexible applications in different scenarios?

In our pilot investigation, we observed that LLMs appear to lack satisfying capabilities in abstract reasoning. As illustrated in Table~\ref{tab:intro}, there is a notable discrepancy—exceeding 17\%—between vanilla accuracy and abstract reasoning accuracy~(\texttt{AbsAcc}) for all LLMs, while humans show only a small disparity. Despite the extensive pre-training of LLMs, which equips them with a vast repository of generic facts, they seem unable to leverage this information as flexibly as humans do~(Figure~\ref{fig:intro}~(b)).

% Parallelly, remarkable headways of large language models have pushed AI much further towards human-like intelligence~\cite{touvron2023llama,yang2023baichuan}. Supervised finetuning~\cite{xu2023wizardlm,mitra2023orca} and prompt engineering~\cite{wei2022chain,kojima2022large} are developed to respectively enhance and stimulate the reasoning abilities of LLMs. Interestingly, can LLMs act like humans to intentionally consider the generic fact behind the problems for solutions development? As shown in Figure~\ref{fig:intro}~(b), LLMs cannot get everything right when facing a series of questions supported by a generic fact. LLMs might exclusively observe and memorize the information that ``hydrochloric acid can dissolve rock'', leading them to overlook the generic fact behind it and hindering their ability to consciously utilize the generic fact for reasoning. This raises doubts that LLMs appear to be rote learning rather than reasoning~\cite{wu2023reasoning}.

In this paper, we aim to systematically and quantitatively investigate the abstract reasoning of LLMs. We first formally define the abstract reasoning metric. Then we carry out abstract reasoning and knowledge probing tasks for a preliminary study. Our investigation comprises two main components: the reasoning tasks can quantify the abstract reasoning capabilities of LLMs. The generic fact probing task can give a further analysis from the perspective of generic fact mastery. Hereafter, we improve abstract reasoning in LLMs based on the analysis of the preliminary study.
Unlike existing chain-of-thought~(CoT) methods~\cite{wei2021finetuned,kojima2022large}, which conduct reasoning with uncontrolled step-by-step explanation, we create a 
dataset AbsR tailored for abstract reasoning. AbsR not only includes generic facts but also their guided explanations, offering a coached approach to understand the process of reasoning with common patterns. Finally, to make LLMs subconsciously exploit the generic fact like humans, we design a simple but effective learning paradigm called meaningful learning~(MeanLearn) to simulate the implicit knowledge learning process~\cite{du2019modeling}, enabling LLMs to implicitly learn and utilize generic facts for reasoning without requiring generic facts as input.

We evaluate MeanLearn on six out-of-domain~(OOD) reasoning and language understanding benchmarks.
Experimental results demonstrate that MeanLearn not only improves general reasoning of LLMs but also excels in abstract reasoning. This distinction in performance is particularly pronounced in the realm of abstract reasoning, underscoring the unique efficacy of our approach in nurturing the higher-order thinking skills that are essential for sophisticated cognitive processing in artificial intelligence systems. Further analysis and ablation studies yield additional evidence supporting the effectiveness of our approach.
We summarize our contributions as follows:
\begin{itemize}
    \item We provide a systematic and quantitative analysis of abstract reasoning in current LLMs, and we develop an abstract reasoning dataset with generic-fact-guided explanations.
    \item We propose meaningful learning to improve abstract reasoning in LLMs with generic fact.
    \item We achieve significant improvement in general reasoning and abstract reasoning on various OOD reasoning and language understanding benchmarks.
\end{itemize}

\begin{wrapfigure}{Rt}{0.53\textwidth}
% \vspace{-1.5cm}
% \vspace{-0.3cm} 
    \centering
    \includegraphics[scale=0.5]{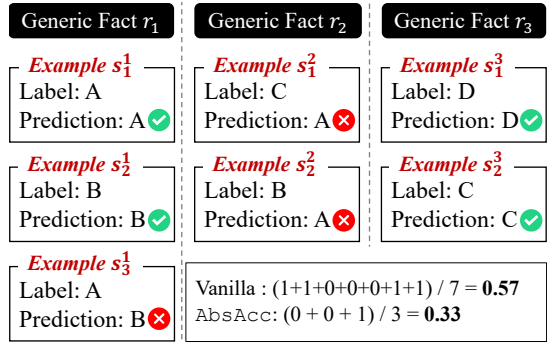}
    \caption{Computation of abstract reasoning metric.}
    \label{fig:metrics}
% \vspace{-1cm}
\vspace{-0.4cm}
\end{wrapfigure}

\section{Abstract reasoning study}
In this section, we initially define the evaluation metric of abstract reasoning to establish quantifiable standards. Then we introduce the LLMs used in this study. Finally, we conduct two experiments: one to directly estimate the abstract reasoning abilities of LLMs, and the other to further analyze it from the perspective of general fact mastery.

\subsection{Abstract reasoning metric}
\label{sec:metric}
% Abstract reasoning is a cognitive process that using the infered general patterns and rules behind samples to genuinely and robustnessly answer question. This form of reasoning transcends specific contexts and can be applied to new problems and situations. 
\textbf{Abstract reasoning} requires models to apply general patterns or high-level abstractions to different scenarios or questions. Different from general reasoning, which focuses on each single example, abstract reasoning takes the examples supported by a generic fact as a whole. We define abstract reasoning accuracy due to the lack of a proper metric.
% in hard constraint and soft contraint.

Given a dataset $D$ with $n$ generic facts $R=\{r_1,r_2\cdots,r_n\}$, for the $i$-th generic fact $r_i$, it supports $m_i$ examples $S_i=\{s^i_1,s^i_2,\cdots,s^i_{m_i}\}\in D$.

As shown in Figure~\ref{fig:metrics}, akin to \citet{qiu2023phenomenal}, we suppose an LLM $\mathcal{LM}$ has grasped the generic fact $r_i$ if and only if $\mathcal{LM}$ can correctly answer all examples in $S_i$. Hence, we define abstract reasoning accuracy~(denoted as \texttt{AbsAcc}) as the proportion of generic facts that $\mathcal{LM}$ has grasped:
\begin{equation}
    \begin{aligned}
        \texttt{AbsAcc} = \frac{\sum_i\mathbb{I}(\Phi(\mathcal{LM}, S_i)=|S_i|)}{n},
    \end{aligned}
\end{equation}
where $\Phi$ quantifies the number of examples in $S_i$ that $\mathcal{LM}$ answers correctly. $\mathbb{I}$ serves as the indicator function, taking 1 when the specified condition is met and 0 otherwise.

% We also consider a soft constraint to have a smoother evaluation of abstract reasoning. Instead of considering the generic rule is not grasped if there is a single mistake in answering $S_i$, we use the ratio of correctly answered samples in $S_i$ to represent the degree of $\mathcal{LM}$ has learned $r_i$:
% \begin{equation}
%     \begin{aligned}
%         \texttt{AbsAcc}\text{-}\texttt{S} = \frac{1}{n}\sum_i\frac{\Phi(L, S_i)}{|S_i|}.
%     \end{aligned}
% \end{equation}

% With \texttt{AbsAcc}-(\texttt{H} and \texttt{S}), we can intuitively understand the abstract performance of current LLMs.

\subsection{Large language models for evaluation}
We employ several LLMs in our preliminary experiments: (1)~\textbf{LLaMA-2}~\cite{touvron2023llama} is an open-access LLM. We use 7B and 13B LLaMA-2 for experiments; (2)~\textbf{Orca-2}~\cite{mitra2023orca} is an open-access LLM finetuned on LLaMA-2 with over 80K reasoning-specific examples. We use 7B and 13B Orca-2 for experiments. (3)~\textbf{GPT-3.5}~\cite{brown2020language} is a limited access LLM, we use \texttt{gpt-3.5-turbo-0125} for experiments.
% \end{itemize}

% \begin{wraptable}{R}{0.35\textwidth}
% \vspace{-1cm}
% \setlength\tabcolsep{2.5pt}
% \caption{The statistics of original and filtered e-CARE.}
% \label{tab:ecare}
% \centering
% \begin{tabular}{cccc}
% \toprule
% \multicolumn{2}{c}{\textbf{Examples}} & \multicolumn{2}{c}{\textbf{Generic Facts}} \\ \midrule
% Original    & Filtered   & Orginal  & Filtered \\ \midrule
% 21,314      & 13,877     & 13,045   & 5,608    \\ \bottomrule
% \end{tabular}
% \end{wraptable}

\begin{table}[t]
    \begin{minipage}[l]{0.42\textwidth}
    \small
        \setlength\tabcolsep{2.2pt}
        \caption{Accuracy of generic fact probing.}
        \vspace{0.2cm}
        \label{tab:gfp}
        \centering
        \begin{tabular}{lccccc}
        \toprule
        \multirow{2}{*}{\textbf{LLMs}} & \multicolumn{2}{c}{\textbf{LLaMA-2}} & \multicolumn{2}{c}{\textbf{Orca-2}} & \textbf{GPT-3.5} \\ \cmidrule{2-6}
                           & 7B      & 13B     & 7B     & 13B  & >20B    \\ \midrule
        \textbf{Accu.} & 62.44   & 70.32   & 26.97  & 12.78 & 77.30  \\ \bottomrule
        \end{tabular}
    \end{minipage}
    \quad
    \begin{minipage}[r]{0.55\textwidth}
        \small
        \setlength\tabcolsep{2.8pt}
        \caption{The categorized abstract reasoning accuracy based on whether the generic facts are known.}
        \vspace{0.2cm}
        \label{tab:cat}
        \centering
        \begin{tabular}{lcccccc}
        \toprule
        \multirow{2}{*}{\textbf{Category}} & \multirow{2}{*}{\textbf{Metrics}} & \multicolumn{2}{c}{\textbf{LLaMA-2}} & \multicolumn{2}{c}{\textbf{Orca-2}} & \textbf{GPT-3.5} \\ \cmidrule{3-7} 
        \multicolumn{2}{l}{}               & 7B    & 13B   & 7B    & 13B & >20B   \\ \midrule
        \textbf{Known}   & \texttt{AbsAcc} & 34.26 & 43.51 & 64.98 & 81.51 & 68.31 \\ 
        \textbf{Unknown} & \texttt{AbsAcc} & 17.19 & 31.84 & 49.00 & 58.38 & 58.49 \\ \bottomrule
        \end{tabular}
    \end{minipage}
\end{table}

% \begin{table}[t]
% \caption{The overall results of the generic facts probing.}
% \label{tab:gfp}
% \centering
% \begin{tabular}{lcccc}
% \toprule
% \multirow{2}{*}{\textbf{LLMs}} & \multicolumn{2}{c}{\textbf{LLaMA-2}} & \multicolumn{2}{c}{\textbf{Orca-2}} \\ \cmidrule{2-5}
%                    & 7B      & 13B     & 7B     & 13B      \\ \midrule
% \textbf{Accuracy } & 62.44   & 98.86   & 26.97  & 12.78    \\ \bottomrule
% \end{tabular}
% \end{table}

% \begin{table}[t]
% \caption{The categorized abstract reasoning accuracy based on whether the generic facts are known.}
% \label{tab:cat}
% \centering
% % \setlength\tabcolsep{5pt}
% \begin{tabular}{lccccc}
% \toprule
% \multirow{2}{*}{\textbf{Category}} & \multirow{2}{*}{\textbf{Metrics}} & \multicolumn{2}{c}{\textbf{LLaMA-2}} & \multicolumn{2}{c}{\textbf{Orca-2}} \\ \cmidrule{3-6} 
% \multicolumn{2}{l}{}               & 7B    & 13B   & 7B    & 13B   \\ \midrule
% \textbf{Known}   & \texttt{AbsAcc} & 34.26 & 43.51 & 64.98 & 81.51 \\ 
% \textbf{Unknown} & \texttt{AbsAcc} & 17.19 & 41.84 & 49.00 & 58.38 \\ \bottomrule
% \end{tabular}
% \end{table}

\subsection{Experiment I: abstract reasoning}
We formulate the first preliminary experiment to gain an initial insight into the abstract reasoning abilities of current LLMs. 

To achieve this goal, we choose a multiple-choice question-answering dataset e-CARE~\cite{du2022care} for implementation. e-CARE is a large-scale explainable causal reasoning dataset with a generic fact in each example, and a generic fact can support more than one example. The full set~(train, test, and dev) of e-CARE is selected for experiments. We filter out the examples that the corresponding generic facts support only one example. Finally, we obtain 13,045 examples supported by 5,608 generic facts. 
% Table~\ref{tab:ecare} shows the statistics of e-CARE.

% \begin{table}[t]
% \centering
% \small
% % \setlength\tabcolsep{5pt}
% \begin{tabular}{llcc}
% \toprule
% \textbf{LLMs}            & \textbf{Size}   & Vanilla Accuracy & \texttt{AbsAcc} \\ \midrule
% \multirow{2}{*}{LLaMA-2} & 7B   &  51.67   &  24.82   \\
%                          & 13B  &  68.45   &  42.94   \\ \midrule
% \multirow{2}{*}{Orca-2}  & 7B   &  74.23   &  51.19   \\
%                          & 13B  &  79.83   &  59.47   \\ \bottomrule
% \end{tabular}
% \caption{The overall reasoning and abstract reasoning performance of LLaMA-2 and Orca-2.}
% \label{tab:pre1}
% \end{table}

We follow zero-shot setting to prompt the LLMs for answer selection from multiple choices, and the generic facts are not provided in the inputs~(refer to Appendix~\ref{app:abs_rea}). Table~\ref{tab:intro} shows the overall results:

(1)~The significant disparity (over 17\%) between the vanilla accuracy and \texttt{AbsAcc} for all LLMs proves two aspects: on the one hand, vanilla accuracy is not ideal for estimating abstract reasoning. On the other hand, although LLMs could answer questions accurately, they still cannot capture the common patterns or generic facts behind the questions. 
% The significant disparity (over 17\%) between vanilla accuracy and \texttt{AbsAcc} for all LLMs shows two things: vanilla accuracy isn't ideal for estimating abstract reasoning, and LLMs, while accurate, can't grasp the common patterns or generic facts behind the questions.
% Hence, the vanilla accuracy does not provide a comprehensive assessment for the reasoning ablility of LLMs. LLMs behave more like imitating or memorizing the training corpus rather than reasoning with their minds.

(2)~Larger LLMs outperform smaller ones due to their richer knowledge and stronger reasoning ability. Orca-2, additional trained on LLaMA-2, notably outperforms LLaMA-2, indicating that supervised fine-tuning enhances both reasoning and abstract reasoning abilities. However, even with large-scale post-training, the significant gap between vanilla accuracy and \texttt{AbsAcc} remains unbridged.

(3)~The minimal gap between vanilla accuracy and \texttt{AbsAcc} for humans suggests the large gap for LLMs is not due to varying difficulties of the example supported by the same generic fact.
% Larger LLMs yield better performance, this is because larger LLMs possess greater and more accurate knowledge. Orca-2, post-trained on LLaMA-2 with enumerous reasoning examples, siginificantly surpasses LLaMA-2, suggesting supervised finetuning improves reasoning and abstract reasoning. Despite this, the large gap between vanilla accuracy and \texttt{AbsAcc}-\texttt{H} cannot be bridged through large-scale and reasoning-specific post-training.

% (3)~\texttt{AbsAcc}-\texttt{S} of all LLMs is close to the vanilla accuracy. When each supported example of a fact is relatively few, \texttt{AbsAcc}-\texttt{S} degenerates into vanilla accuracy.
% (3)~\texttt{AbsAcc}-\texttt{S} of all LLMs is close to the vanilla accuracy, this is due to .

\subsection{Experiment II: generic fact probing}
To delve into the abstract reasoning from the perspective of generic fact mastery, we design a generic fact probing task to ask LLMs whether they possess the given generic fact~(refer to Appendix~\ref{app:fact_probing} for more details). Drawing inspiration from knowledge uncertainty estimation~\cite{amayuelas2023knowledge}, we formulate this task as a yes or no problem. The overall results are shown in Table~\ref{tab:gfp}. We could conclude as follows:

(1)~LLaMA-2 and GPT-3.5 have substantial knowledge reserves. This indicates the poor \texttt{AbsAcc} of LLaMA-2 might come from a lack of ability to apply the generic fact to different reasoning scenarios. This problem also exists in GPT-3.5, but it is not as serious as LLaMA-2.

(2)~In contrast, Orca-2 knows limited generic facts, and the larger the model, the less it knows. This contrasts sharply with its strong reasoning abilities~(vanilla accuracy). We suppose training Orca-2 is more like rote learning rather than meaningful learning, which implies that Orca-2 tends to lose knowledge while acquiring reasoning.

% \subsection{Further Analysis}
\label{sec:2-5}
To deeply investigate the relation between generic facts and abstract reasoning, we categorize the examples in e-CARE according to whether LLMs know the generic facts. The results are represented in Table~\ref{tab:cat}, from which we can draw the following conclusions:

(1)~The performance of ``Known'' category has superiority over ``Unknown'' category across all LLMs. This demonstrates knowing the generic facts is helpful for abstract reasoning, which provides a potential avenue for abstract reasoning enhancement through generic fact retrieval.

(2)~Nonetheless, most of the results in the ``Unknown'' category still have undeniable performance, suggesting that LLMs may engage in reasoning based on suspicious correlations or rely on memorization. 
% We can reasonably speculate that LLMs may consciously conduct memorization when answering reasoning-related questions.

In conclusion, we can enhance the abstract reasoning abilities of LLMs in two ways: (1)~\textbf{\textit{Knowledge}}, injecting generic facts through additional training or prompting; (2)~\textbf{\textit{Reasoning}}, teaching LLMs how to utilize the generic facts to provide more precise responses to questions.

\begin{figure*}
    \centering
    \includegraphics[width=1\textwidth]{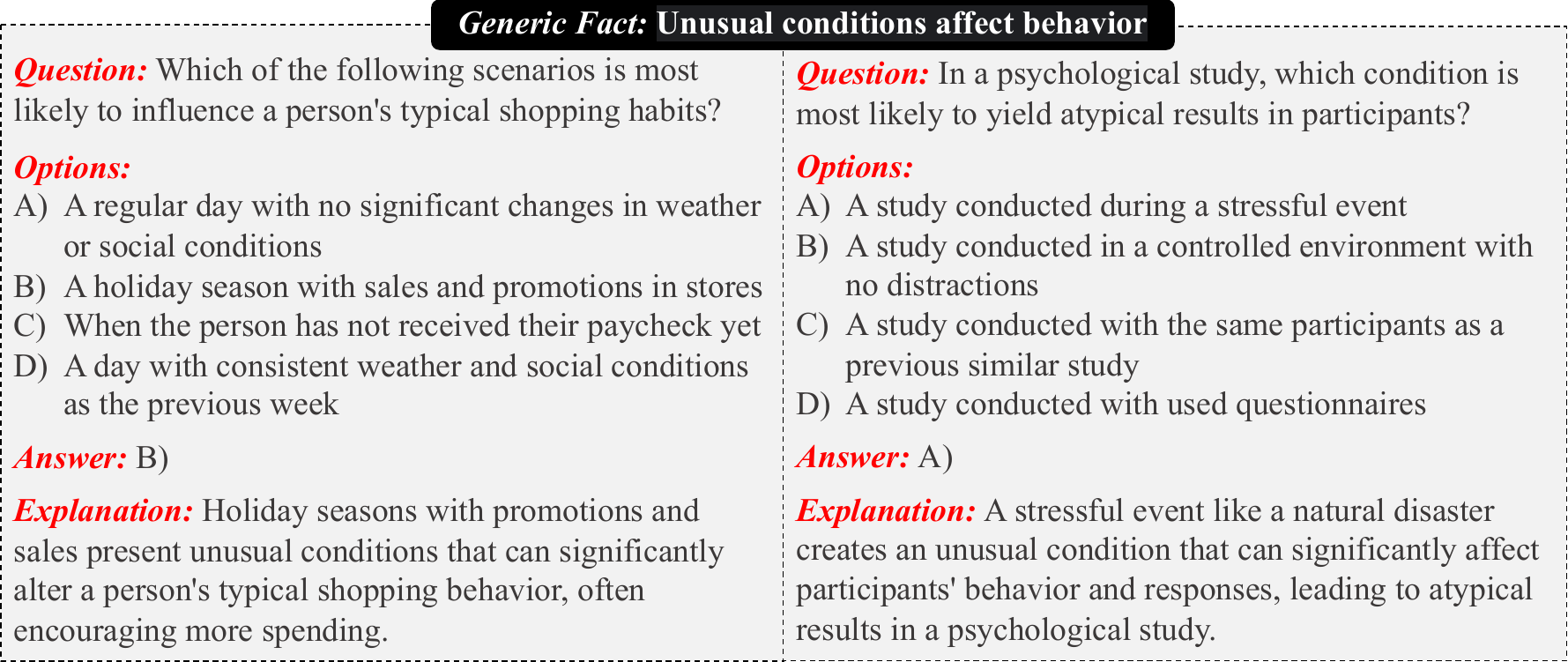}
    \caption{Two samples of AbsR guided by the generic fact about ``unusual conditions''.}
    \label{fig:example}
\end{figure*}

\section{AbsR: abstract reasoning dataset}
To enhance the abstract reasoning abilities of LLMs through \textbf{\textit{Knowledge}} and \textbf{\textit{Reasoning}} ways, we need a dataset that can not only provide generic facts but also teach LLMs how to exploit the generic fact to derive the correct answer in different scenarios~(meaningful learning). Since there is no suitable dataset to assist us in achieving the goals, we create an \textbf{Abs}tract \textbf{R}easoning dataset~(AbsR).

\subsection{Generic fact collection}
To obtain the generic facts, We chose GenericsKB~\cite{bhakthavatsalam2020genericskb} as the foundational generic fact base for AbsR. GenericsKB is a large-scale knowledge base with over 3.4 million sentences~(e.g. Dogs bark) expressing general truths, each of which also includes the corresponding concept~(e.g. Dog) and the confidence score~(between 0 and 1). Hence, we filter and then sample a collection of 4,613 high-quality generic facts of different concepts from GenericsKB~(details of generic fact filtering and sampling can refer to Appendix~\ref{details:fact_filtering}).
% We apply the following steps to filter generics fact from GenericsKB:

% (1)~Exclude sentences with confidence $<$ 0.7.

% (2)~Categorize the sentences by the concepts.

% (3)~Randomly sample 4,613 concepts, and then randomly sample one sentence as the generic fact from the category of each sampled concept.

% Hereafter, we can obtain a collection of 4,613 generic facts. 

\begin{wraptable}{R}{0.5\textwidth}
\setlength\tabcolsep{3pt}
\vspace{-1.4cm}
\caption{The statistics of AbsR.}
\label{tab:absr}
\centering
\begin{tabular}{lccc}
\toprule
      & \textbf{Examples} & \textbf{Questions} & \textbf{Generic Facts} \\ \midrule
Train & 18,020   & 9,010    & 4,613         \\
Test  & 200      & 200       & 104    \\ \bottomrule      
\end{tabular}
\vspace{-0.5cm}
\end{wraptable}

\subsection{Dataset construction}
\label{sec3.1}

To construct the whole dataset, and inspired by previous works~\cite{wang2022self,xu2023wizardlm}, we choose GPT-4~\cite{achiam2023gpt} as our data annotator. The API we used is \texttt{gpt-4-1106-preview}.

Specifically, for each sampled generic fact $r_i$, we would ask GPT-4 to create samples $S_i=\{s_1^i,\cdots,s_{m_i}^i| 1\leq m_i\leq 3\}$ in various scenarios based on $r_i$. Each sample $s^i_j$ contains the following terms: (1) a question $X^i_j$ with a few options, (2) a response $Y^i_j$ with an answer and an explanation guided by $r_i$. All terms form a triple $s^i_j=<X^i_j, r_i, Y^i_j>$ for each sample. The prompt for dataset creation can refer to Appendix~\ref{pmp:dataset_construction}. Figure~\ref{fig:example} shows an illustration of the created samples.

Finally, without loss of generality, given the $j$-th sample $s^i_j$ of generic fact $r_i$, we can create two kinds of examples for meaningful learning. One is predicting $Y^i_j$ given $<X^i_j, r_i>$~(K-example), while the other is predicting $Y^i_j$ given only $X^i_j$~(R-example). The examples can implicitly enhance abstract reasoning in LLMs through \textbf{\textit{Knowledge}} and \textbf{\textit{Reasoning}} ways. Table~\ref{tab:absr} shows the statistics of AbsR.

\subsection{The quality of AbsR}
We conduct human evaluations to measure the quality of AbsR from the following dimensions~(details and more statistics such as evaluation criteria, agreements, and pay can refer to Appendix~\ref{app:human} and~\ref{app:instruction}):
\begin{itemize}
    \item \textbf{Human Performance}: humans can achieve vanilla accuracy of 95\% and \texttt{AbsAcc} of 93.27\%;
    \item \textbf{Support Rate}: 89\%~(the rate of examples which can be supported by the generic fact);
    \item \textbf{Diversity}: 88.5\%~(the diversity of the examples supported by the same generic fact. Greater diversity indicates greater differences among the examples of the same generic fact).
\end{itemize}
For comparison, e-CARE~\cite{du2022care} is also human-annotated, with examples generated from generic facts. On e-CARE, humans reached 92\% vanilla accuracy, 89.9\% \texttt{AbsAcc}, and an 87\% support rate~\cite{du2022care}, showing that AbsR matches the quality of the human-annotated dataset. 
% Additionally, the minimal gap between vanilla accuracy and \texttt{AbsAcc} suggests that the large gap between them~(Table~\ref{tab:intro}) is not due to varying difficulties of the example supported by the same generic fact.

\section{Method: meaningful learning}
To imitate humans' instinctive use of general facts in reasoning, we develop a simple but effective learning paradigm called meaningful learning~(MeanLearn). It can enhance the abstract reasoning abilities of LLMs in \textbf{\textit{Knowledge}} and \textbf{\textit{Reasoning}} ways. This is inspired by the implicit knowledge learning process, which uses hidden variables to learn event background knowledge~\cite{du2019modeling}. MeanLearn can make LLMs implicitly learn generic facts and solve problems under the guidance of generic facts. 
% It simulates a process of conditional variational autoencoder~\cite{sohn2015learning}.

Specifically, an LLM $\mathcal{LM}$ with parameters $\theta$ and the input $x$ can construct a conditional probability for the output $y$:
\begin{equation}
    \mathcal{LM}(x,y,\theta)=-\sum_t\log p_\theta(y_t|x,y_{<t}).
\end{equation}
In MeanLearn, for a K-example $<X,r,Y>$ and R-example $<X,Y>$ pair guided by generic fact $r$, we send the example pair into $\mathcal{LM}$ to model two conditional probabilities: 
\begin{equation}
    \begin{aligned}
        \mathcal{LM}(X,Y,\theta)&=-\sum_t\log p_\theta(Y_t|X,Y_{<t}), \\
        \mathcal{LM}(X,r,Y,\theta)&=-\sum_t\log q_\theta(Y_t|X,r,Y_{<t}).
    \end{aligned}
\end{equation}
Hereafter, on the one hand, we train $\mathcal{LM}$ to learn policies $p_\theta(Y_t|X,Y_{<t})$ and $q_\theta(Y_t|X,r,Y_{<t})$ together to enable $\mathcal{LM}$ implicitly learn the generac fact $r$. On the other hand, $\mathcal{LM}$ can learn how to apply $r$ into different scenarios by learning the explanations in $Y$ of different $X$ supported by $r$. Finally, we can reason with the implicit guidance of $r$ when only given $X$, just like humans instinctively answer questions without explicitly giving the general facts.

\section{Experiments}
\label{sec:experiments}
\subsection{Experimental details}
% \subsection{Training Details}
\textbf{Training setup.}\quad We use LoRA~\cite{hu2022lora} for parameter effcient finetuning. 7B and 13B MeanLearn are trained on 7B and 13B Orca-2, respectively, 8B MeanLearn is trained on 8B LLaMA-3, with batch sizes of 256 for 7B and 8B, and 240 for 13B. MeanLearn of various sizes are trained for one epoch at a 5e-5 learning rate with AdamW~\cite{kingma2014adam}. 7B and 8B MeanLearn are trained on 2 NVIDIA A100 80GB PCIe GPUs for 3 hours. 13B MearnLearn is trained on 3 such GPUs for 4 hours.

% of Orca-2 and LLaMA-3.  We train them for one epoch with a learning rate of 5e-5. 7B and 13B MeanLearn are respectively trained on 2 and 3 NVIDIA A100 80GB PCIe GPUs with about 3 hours. MeanLearn-8B is trained on LLaMA-3-8B on 2 NVIDIA A100 80GB PCIe GPUs with about 3 hours.

\textbf{Baselines.}\quad We adopt a wide range of open and limited access LLMs across different sizes as baselines. \textbf{Open Access LLMs}: (1)~\underline{LLaMA-2}~\cite{touvron2023llama} 7B and 13B; (2)~\underline{LLaMA-3}~\cite{meta2024introducing} 8B; (3)~\underline{Vicuna}~\cite{vicuna2023} 7B and 13B~(finetuned on ShareGPT~\cite{sharegpt} data); (4)~\underline{WizardLM}~\cite{xu2023wizardlm} 7B and 13B~(fintuned with evolved instruction data); (5)~\underline{Orca-2}~\cite{mitra2023orca} 7B and 13B~(prograssively finetuned on massive reasoning data). \textbf{Limited Access LLMs}: (1)~\underline{GPT-3.5}~(\texttt{gpt-3.5-turbo-0125}); (2)~\underline{PaLM-2}~(\texttt{text-bison-001}).

\textbf{Evaluation benchmarks.}\quad In addition to AbsR, we also include a wide range of benchmarks of reasoning and natural language understanding for evaluation: (1)~\underline{AGIEval}~\cite{zhong2023agieval} consists of tests ranging from college admission tests, to national civil service examinations; (2)~\underline{RACE}~\cite{lai2017race} consists of tests with reading comprehension questions; (3)~\underline{BBH}~\cite{suzgun-etal-2023-challenging} is a subset of Big-Bench~\cite{srivastava2023beyond}, which contains 23 hardest tasks focusing on challenging scenarios; (4)~\underline{Com.}~\cite{xiong2023examining} is a collection of 7 commonsense reasoning datasets~($\alpha$NLI~\cite{bhagavatula2019abductive}, CSQA~\cite{talmor2019commonsenseqa}, COPA~\cite{roemmele2011choice}, e-CARE~\cite{du2022care}, SocialIQa~\cite{sap2019social}, PIQA~\cite{bisk2020piqa}, and StrategyQA~\cite{geva2021did}); (5)~\underline{MMLU}~\cite{hendrycks2021measuring} is a massive multitask language understanding benchmark; (6)~\underline{ARC}~\cite{clark2018think} is a benchmark of easy~(ARC-e) and challenge~(ARC-c) science questions.

\textbf{Evaluation setup.}\quad Since generation-based evaluation is time-consuming due to limited computing resources, we follow OpenCompass~\cite{2023opencompass} to have hybrid evaluation criteria with high reproducibility: (1)~For PaLM-2 and GPT-3.5 on all tasks, and open access LLMs on 4 generation tasks~(4 tasks in BBH), greedy decoding and exact match are utilized for evaluation; (2)~For open access LLMs on classification and multiple-choice tasks, perplexity~(PPL) is adopted for prediction. The option or category with the lowest PPL is chosen as the answer. For all baselines, evaluation is conducted once since there is no randomness on such evaluation criteria. For MeanLearn, results are averaged of MeanLearn(s) trained with 3 different random seeds. The evaluation prompts can refer to Appendix~\ref{app:evaluation}.

% Take Orca-2 and MeanLearn as an example, we use:

% \begin{wrapfigure}{r}{0.5\textwidth}
% \begin{tcolorbox}[fontupper = \ttfamily, top = 0mm, bottom = 0mm, left = 2mm]
% <|im\_start|>\mybox[green]{system}\\
% \mybox[green]{\{INSTRUCTION\}}<|im\_end|>\\
% <|im\_start|>\mybox[yellow]{user}\\
% \mybox[yellow]{Question:\{QUESTION\}}<|im\_end|>\\
% <|im\_start|>\mybox[red]{assistant}\\
% \mybox[red]{Answer:\{OPTION\}}<|im\_end|>
% \end{tcolorbox}
% \end{wrapfigure}

% The text enclosed within curly braces is a placeholder. \mybox[green]{\{INSTRUCTION\}} will be filled with task-specific instruction. \mybox[yellow]{\{QUESTION\}} will be filled with the question and all candidate options or categories. \mybox[red]{\{OPTION\}} will be filled with the option index only~(e.g. A).

\begin{table}[t]
\caption{Overall vanilla accuracy and \texttt{AbsAcc}~(\%) of baselines and MeanLearn. Due to the space limit, we only report the standard deviation of \textbf{\textit{Average}} performance of each method.}
\vspace{0.2cm}
\label{tab:main}
\small
\centering
\setlength\tabcolsep{4pt}
\small
\begin{tabular}{llccccccccc}
\toprule
\textbf{Size} & \textbf{LLMs} & \textbf{AbsR}  & \textbf{AGIEval} & \textbf{RACE}  & \textbf{BBH}   & \textbf{Com.} & \textbf{MMLU}  & \textbf{ARC-e} & \textbf{ARC-c} & \textbf{\textit{Average}} \\ \midrule
\multicolumn{11}{l}{\mybox[green]{\footnotesize Vanilla Accuracy}}    \\
\multirow{2}{*}{>20B} & PaLM-2    & 75.76  & 15.83 & 75.82 & 48.73 & 78.05 & 58.73 & 89.95 & 82.37 & 65.77\scriptsize{ $\pm$ 0.00} \\
                      & GPT-3.5   & 84.00  & 25.84 & 83.36 & 54.15 & 74.80 & 65.98 & 94.71 & 88.81 & 71.46\scriptsize{ $\pm$ 0.00} \\ \cmidrule{2-11}

\multirow{5}{*}{7B}   & LLaMA-2   & 50.00  & 27.89 & 38.85 & 26.00 & 55.37 & 41.65 & 58.73 & 43.05 & 42.69\scriptsize{ $\pm$ 0.00} \\
                      & Vicuna & 75.00 & 32.56 & 65.75 & 31.93 & 64.77 & 49.40 & 74.78 & 57.63 & 56.48\scriptsize{ $\pm$ 0.00} \\
                      & WizardLM & 65.00 & 22.27 & 22.11 & 25.99 & 43.63 & 23.26 & 25.57 & 22.37 & 31.28\scriptsize{ $\pm$ 0.00} \\
                      & Orca-2    & 73.50  & \textbf{38.57} & 73.32 & 34.51 & 71.58 & 50.11 & 79.19 & 73.90 & 61.84\scriptsize{ $\pm$ 0.00} \\
                      & MeanLearn & \textbf{77.00}  & 38.15 & \textbf{77.15} & \textbf{35.64} & \textbf{76.59} & \textbf{52.98} & \textbf{86.67} & \textbf{78.06} & \textbf{65.28}\scriptsize{ $\pm$ 0.34} \\ \cmidrule{2-11}

\multirow{2}{*}{8B}  & LLaMA-3  & 82.50 & 36.33 & 76.75 & 36.60 & \textbf{71.76} & 61.86 & 88.54 & 75.25 & 66.20\scriptsize{ $\pm$ 0.00} \\
                     & MeanLearn & \textbf{84.50} & \textbf{38.53} & \textbf{77.02} & \textbf{37.67} & 71.73 & \textbf{62.35} & \textbf{88.89} & \textbf{77.97} & \textbf{67.12}\scriptsize{ $\pm$ 0.18} \\ \cmidrule{2-11}

\multirow{5}{*}{13B}  & LLaMA-2   & \textbf{73.50}  & 34.41 & 59.46 & 30.61 & 61.00 & \textbf{51.87} & 71.25 & 55.25 & 54.67\scriptsize{ $\pm$ 0.00} \\
                      & Vicuna & 74.00 & 33.23 & 63.54 & 33.22 & 63.10 & 49.34 & 77.78 & 59.77 & 56.75\scriptsize{ $\pm$ 0.00} \\
                      & WizardLM & 74.00 & 33.23 & 63.54 & 33.22 & 63.10 & 49.34 & 77.78 & 59.77 & 56.75\scriptsize{ $\pm$ 0.00} \\
                      & Orca-2    & 66.50  & 43.75 & 66.86 & 39.39 & 65.92 & 47.27 & 82.19 & 70.85 & 60.34\scriptsize{ $\pm$ 0.00} \\
                      & MeanLearn & 67.17  & \textbf{43.92} & \textbf{70.24} & \textbf{40.36} & \textbf{69.75} & 51.68 & \textbf{88.69} & \textbf{82.00} & \textbf{64.23}\scriptsize{ $\pm$ 0.25} \\ \midrule 
\multicolumn{11}{l}{\mybox[red]{AbsAcc}}     \\
\multirow{2}{*}{>20B} & PaLM-2    & 64.58 & 9.59 & 61.54 & 27.69 & 69.34 & 44.39 & 85.68 & 77.10 & 54.99\scriptsize{ $\pm$ 0.00} \\
                      & GPT-3.5   & 77.08 & 14.48 & 66.79 & 25.45 & 65.74 & 52.09 & 92.16 & 85.05 & 59.86\scriptsize{ $\pm$ 0.00} \\ \cmidrule{2-11}

\multirow{5}{*}{7B}   & LLaMA-2   & 35.42 & 16.09  & 14.08  & 9.56  & 47.42  & 25.48  & 50.27  & 35.51  & 29.23\scriptsize{ $\pm$ 0.00} \\
                      & Vicuna    & 58.33 & 19.90 & 38.68 & 14.28 & 58.07 & 32.50 & 65.95 & 49.53 & 39.84\scriptsize{ $\pm$ 0.00} \\
                      & WizardLM & 53.24 & 12.94 & 5.57 & 9.77 & 32.96 & 11.55 & 20.00 & 17.29 & 20.42\scriptsize{ $\pm$ 0.00} \\
                      & Orca-2    & 60.42 & \textbf{26.27}  & 50.90 & 15.67  & 64.02 & 33.68  & 72.16 & 67.76 & 48.86\scriptsize{ $\pm$ 0.00} \\
                      & MeanLearn & \textbf{64.58} & 25.27  & \textbf{57.10} & \textbf{16.67}  & \textbf{71.17} & \textbf{37.24}  & \textbf{81.35} & \textbf{73.83} & \textbf{53.39}\scriptsize{ $\pm$ 0.32} \\ \cmidrule{2-11}

\multirow{2}{*}{8B}  & LLaMA-3  & 72.92 & 23.78 & \textbf{55.61} & 18.13 & 66.35 & \textbf{47.31} & 83.51 & 70.09 & 54.71\scriptsize{ $\pm$ 0.00} \\
                     & MeanLearn & \textbf{73.96} & \textbf{26.57} & 55.09 & \textbf{19.24} & \textbf{66.48} & 46.76 & \textbf{83.78} & \textbf{73.36} & \textbf{55.66}\scriptsize{ $\pm$ 0.17} \\ \cmidrule{2-11}

\multirow{5}{*}{13B}  & LLaMA-2   & \textbf{61.46} & 21.01  & 32.17  & 13.74  & 52.87  & \textbf{35.72}  & 62.43 & 44.86  & 40.53\scriptsize{ $\pm$ 0.00} \\
                      & Vicuna    & 61.46 & 19.70 & 36.50 & 16.33 & 55.26 
                      & 32.21 & 70.27 & 52.34 & 40.37\scriptsize{ $\pm$ 0.00} \\
                      & WizardLM & 59.38 & 16.96 & 28.59 & 18.61 & 55.73 & 29.47 & 57.84 & 42.06 & 38.58\scriptsize{ $\pm$ 0.00} \\
                      & Orca-2    & 50.00 & 29.50  & 42.12 & 20.61  & 56.41  & 31.11  & 75.95 & 62.15 & 45.98\scriptsize{ $\pm$ 0.00} \\
                      & MeanLearn & 45.82 & \textbf{30.01}  & \textbf{47.65} & \textbf{21.85}  & \textbf{61.04} & 35.21  & \textbf{85.41} & \textbf{77.08} & \textbf{50.51}\scriptsize{ $\pm$ 0.23} \\ \bottomrule
\end{tabular}
\end{table}

\subsection{Results: vanilla accuracy}
Table~\ref{tab:main} shows the \mybox[green]{Vanilla Accuracy} of MeanLearn and baselines:

(1)~Additional training usually brings improvements ~(Vicuna, WizardLM, Orca-2, and MeanLearn), but it still can result in performance losses on some benchmarks~(e.g. AGIEval). This signifies the effectiveness of massive reasoning-specific data and the necessity of data coverage. 

(2)~MeanLearn has superiority over all open access baselines on nearly all benchmarks. This demonstrates that MeanLearn can use generic facts to effectively guide LLMs to reason more properly and logically, making them more flexible and task-adaptive.

(3)~The more challenging the benchmark~(e.g. BBH), the less pronounced the improvement in the performance of MeanLearn. This is mainly because the improvement is mostly constrained by the complex scenarios in harder benchmarks, which require mightier base LLMs or more complex training data to perform and learn multi-step and compositional reasoning.

(4)~MeanLearn does not have many advantages over LLaMA-3. LLaMA-3 stores dense knowledge in its parameters, making it harder to learn new knowledge without forgetting existing knowledge.
% (4)~Post-training~(WizardLM, Vicuna, Orca-2 and MeanLearn) does not always bring improvements~(e.g. AGIEval). The primary reason is the reasoning-specific post-training, which is bound to affect the general abilities and the handle of uncovered reasoning types.

(5)~Larger open-sourced LLMs do not always yield better results. We suppose the main factors behind this are in two ways: on the one hand, LLMs with different parameter sizes might be good at different tasks, and excessive contemplation may lead to decreases on some tasks~\cite{wei2022chain}. On the other hand, the perplexity-based evaluation setting might have moderate cross-scale generalization ability. 

(6)~LLMs possess more than 20B parameters still struggle with abstract reasoning. MeanLearn can achieve comparable performance to PaLM-2 despite their huge gap in model sizes.

\begin{wrapfigure}{R}{0.515\textwidth}
    \vspace{-0.5cm}
    \centering
    \includegraphics[width=0.51\textwidth]{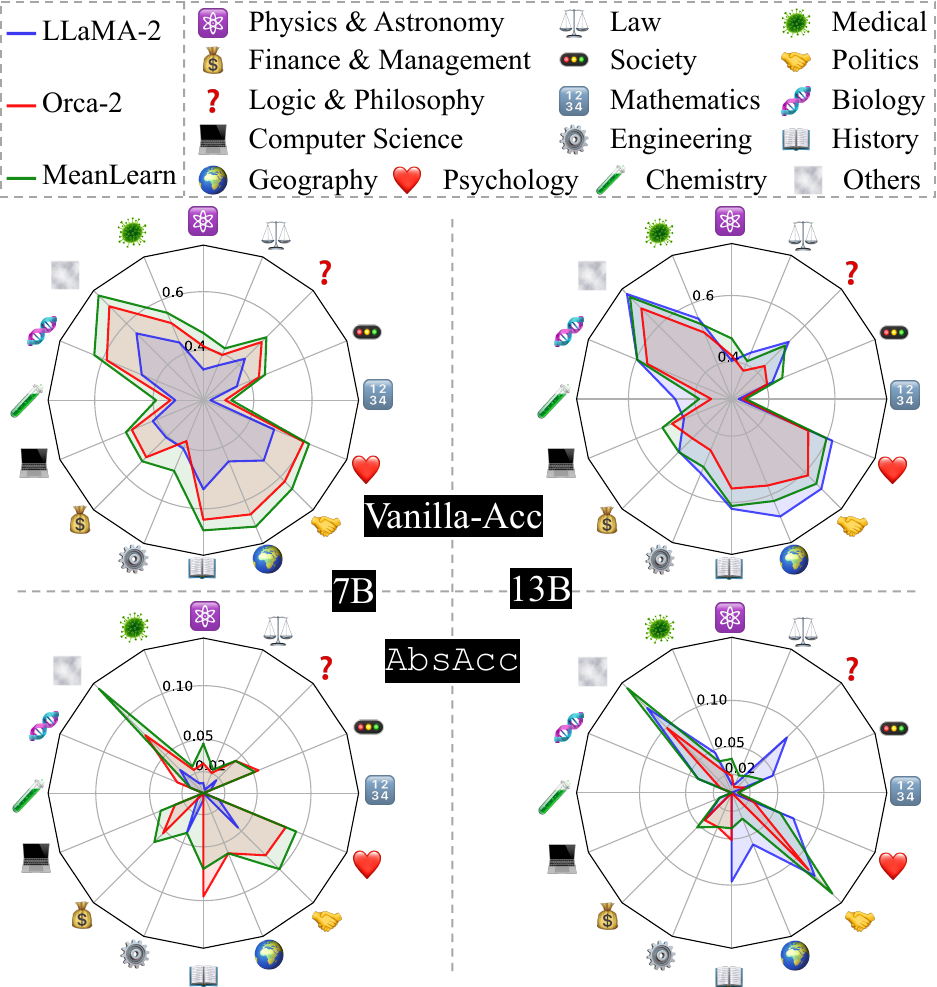}
    \caption{Visualization of reasoning capabilities on MMLU, which is categorized by task domians.}
    \label{fig:domain}
    \vspace{-1.5cm}
\end{wrapfigure}

\subsection{Results: abstract reasoning accuracy}
Since the generic facts are not given in the OOD benchmarks, we train a RoBERTa-Large~\cite{liu2019roberta} with constrastive learning to cluster examples supported by the same generic fact. In particular, we utilize e-CARE as the training set, examples guided by the same generic fact will be pushed closer, or they will be pushed away. Cosine similarity is used to measure the distance between examples~(training details can refer to Appendix~\ref{app:roberta}). For clustering, examples sharing a similarity above 0.6 are considered to be supported by the same generic fact. Each cluster contains no more than 3 examples.

Hereafter, we calculate the abstract reasoning metrics based on the results of clustering. The overall results are shown in Table~\ref{tab:main}~(\mybox[red]{AbsAcc}), we can have the following observations:

(1)~\texttt{AbsAcc} is much lower than vanilla accuracy, indicating the large gap between general reasoning and abstract reasoning. More efforts should be made to fill this gap in the future.
% Perhaps more training data can help to fill the gap between vanilla accuracy and abstract reasoning accuracy.

% (2)~Nevertheless, MeanLearn outperforms Orca-2 and LLaMA-2 on \texttt{AbsAcc}. This reveals the significance of MeanLearn in enhancing abstract reasoning via meaningful learning.

% (3)~Advantages on vanilla accuracy does not always lead to advantages on \texttt{AbsAcc}~(AGIEval). This spotlights the necessity of using proprietary evaluation metrics to evaluate the abstract reasoning abilities of LLMs.

(2)~MeanLearn can surpass all baselines, usually exhibiting greater advantages in \texttt{AbsAcc} than in vanilla accuracy. This suggests that the improvement in vanilla accuracy is more likely due to the enhancement of abstract reasoning abilities. MeanLearn can tell LLMs the generic facts and teach them how to implicitly utilize them in different scenarios, resulting in better abstract reasoning.

% (4)~7B LLMs seems to outperform 13B LLMs. We suppose this is mainly due to large LLMs tend to prefer generating explanations before generating answers. Our evaluation strategy that directly gives answer might not be favorable to larger LLMs.

\section{Further analysis}
To further investigate the pros, cons, and effectiveness of MeanLearn, we conduct several analyses and ablation studies. LLaMA-2, Orca-2, and MeanLearn are adopted for analysis.
% We choose vanilla accuracy and \texttt{AbsAcc-H} as the evaluation metrics.

\begin{table}[t]
\caption{The overall reasults of ablation studies. ``w/o'' denotes without.}
\vspace{0.2cm}
\label{tab:ablation}
\small
\centering
\setlength\tabcolsep{4pt}
\begin{tabular}{llccccccccc}
\toprule
\textbf{Size} & \textbf{Method} & \textbf{AbsR} & \textbf{AGIEval} & \textbf{RACE}  & \textbf{BBH}   & \textbf{Com.} & \textbf{MMLU}  & \textbf{ARC-e} & \textbf{ARC-c} & \textbf{\textit{Average}} \\ \midrule
\multicolumn{10}{l}{\mybox[green]{Vanilla Accuracy}}                                                                                                   \\
\multirow{4}{*}{7B}   & MeanLearn  & \textbf{77.00}  & 38.15 & \textbf{77.15} & \textbf{35.64} & \textbf{76.59} & \textbf{52.98} & \textbf{86.67} & \textbf{78.06} & \textbf{65.28}       \\
                      & w/o Knowledge  & 73.50 & 38.89   & 74.10 & 33.80 & 73.28       & 50.97 & 79.72 & 75.25 & 62.44       \\
                      & w/o Reasoning  & 72.50 & 38.37   & 72.60 & 33.39 & 71.68       & 49.55 & 77.25 & 73.22 & 61.07       \\
                      & w/ AbsR$^*$   & 73.50 & \textbf{38.53}   & 73.14 & 33.65 & 73.57       & 50.96 & 80.60 & 75.25 & 62.40       \\ \cmidrule{2-11} 
\multirow{4}{*}{13B}  & MeanLearn   & \textbf{67.17}  & 43.92 & \textbf{70.24} & \textbf{40.36} & 69.75 & \textbf{51.68} & \textbf{88.69} & \textbf{82.00} & \textbf{64.23}  \\
                      & w/o Knowledge  & 63.50 & \textbf{44.07}   & 67.19 & 36.80 & \textbf{70.45}   & 47.88 & 84.13 & 74.58 & 61.08       \\
                      & w/o Reasoning  & 56.00 & 43.95   & 58.05 & 35.91 & 68.09       & 45.31 & 82.54 & 69.83 & 57.46       \\
                      & w/ AbsR$^*$  & 66.50 & 43.75   & 66.86 & 36.92 & 65.92       & 47.27 & 82.19 & 70.85 & 60.03       \\ \cmidrule{2-11} 
\multicolumn{10}{l}{\mybox[red]{AbsAcc}}                                                                                                           \\
\multirow{4}{*}{7B}   & MeanLearn  & \textbf{64.58} & 25.27  & \textbf{57.10} & \textbf{16.67}  & \textbf{71.17} & \textbf{37.24}  & \textbf{81.35} & \textbf{73.83} & \textbf{53.39}       \\
                      & w/o Knowledge & 61.46  & \textbf{26.67}    & 52.91  & 14.74   & 66.26  & 34.84 & 72.70 & 69.16 & 49.84       \\
                      & w/o Reasoning & 59.38  & 26.18 & 49.66 & 14.15  & 64.13   & 33.27  & 69.19 & 67.29  & 47.91  \\
                      & w/ AbsR$^*$  & 61.46 & 26.44 & 52.24 & 14.76  & 55.56   & 34.88  & 74.05 & 69.16 & 48.57  \\ \cmidrule{2-11} 
\multirow{4}{*}{13B}  & MeanLearn  & 45.82 & 30.01  & \textbf{47.65} & \textbf{21.85}  & 61.04 & 35.21  & \textbf{85.41} & \textbf{77.08} & \textbf{50.51}       \\
                      & w/o Knowledge & 46.88 & 30.43    & 42.87 & 18.83  & \textbf{62.95} & 32.10  & 78.11 & 66.82 & 47.37       \\
                      & w/o Reasoning & 36.46  & \textbf{30.81}    & 31.19  & 18.26  & 59.70 & 29.12  & 75.68 & 61.68 & 42.86        \\
                      & w/ AbsR$^*$  & \textbf{50.00} & 29.58    & 42.12 & 18.82  & 56.41  & 31.11  & 75.95 & 62.15 & 45.77        \\ \bottomrule
\end{tabular}
\end{table}

\subsection{Performance on different domains}
\label{sec: visualization}
We categorize tasks of MMLU based on their respective domains to demonstrate the pros and cons of MeanLearn for future improvements. The results are shown in Figure~\ref{fig:domain} with a total of 16 categories:

(1)~All LLMs possess poor vanilla accuracy and \texttt{AbsAcc} in math and chemistry. We conjecture the primary factors are the complexity of math problems and the shortage of chemistry knowledge.

% (2)~For all LLMs, there is still a noticeable performance gap between vanilla accuracy and \texttt{AbsAcc-H} in geography, medical, and biology domains. This implies that LLMs are more likely memorizing rather than reasoning in these domains.
(2)~Compared to Orca-2, MeanLearn possesses considerable advantages in engineering, geography, and physics \& astronomy domains. We suppose this is related to the domain coverage of AbsR.

(3)~Since the overall \texttt{AbsAcc} is not relatively high, for potential improvement in the future, apart from focusing more on the domains with poor vanilla accuracy and \texttt{AbsAcc}, we should also increase the overall volume of finetuning data.

\subsection{How much do knowledge and reasoning ways contribute to MeanLearn?}
We utilize \textbf{\textit{Knowledge}} and \textbf{\textit{Reasoning}} ways to enhance abstract reasoning in LLMs. To investigate the effect of them, we conduct an ablation study of AbsR to respectively remove the \textbf{\textit{Knowledge}}~(generic facts) and \textbf{\textit{Reasoning}}~(explanations) in each example. Results are shown in Table~\ref{tab:ablation}~(w/o Knowledge and w/o Reasoning), we can observe:

(1)~Removing \textbf{\textit{Knowledge}}~(generic facts) or \textbf{\textit{Reasoning}}~(explanation) leads to a decrease in performance. This is because the generic facts and their guided reasoning process in AbsR can teach LLMs to learn and utilize the generic fact behind the questions to reason more logically and properly.

% (2)~MeanLearn also obtains a drop in performance while eliminating \textbf{\textit{Reasoning}}~(explanation) in each example of AbsR. This indicates the generic fact guided reasoning process can teach LLMs to reason more logically and properly.

(2)~In MeanLearn, \textbf{\textit{Knowledge}} contributes less in boosting the vanilla and abstract reasoning than \textbf{\textit{Reasoning}}. It indicates the importance of teaching LLMs how to reason under the guidance of generic facts. Furthermore, it reveals the bottleneck of small-scale LLMs reasoning is caused more by reasoning itself rather than knowledge.

(3)~MeanLearn without Knowledge can defeat Orca-2 on vanilla accuracy and \texttt{AbsAcc}, while MeanLearn without Reasoning cannot. This implies the synergy effects between \textbf{\textit{Knowledge}} and \textbf{\textit{Reasoning}}~(MeanLearn) can push LLMs further.

(4)~In AGIEval, \textbf{\textit{Knowledge}} and \textbf{\textit{Reasoning}} are less critical due to their focus on complex reading comprehension without needing vast knowledge. Moreover, \textbf{\textit{Knowledge}} seems useless in Commonsense. When providing LLMs with knowledge, we should also teach them how to use it.

\subsection{Is the improvement in performance solely attributed to additional data?}
Supervised fine-tuning can also improve \texttt{AbsAcc}~(Table~\ref{tab:intro}), so we design another ablation study to investigate whether the improvement of MeanLearn only comes from additional data.
% When constructing AbsR, we would ask GPT-4 to generate an explanation guided by the generic fact~(Sec.~\ref{sec3.1}). Hence, we devise another ablation study to investigate the effect of the guidance brought by the generic fact.

To be precise, for each sample in AbsR, we use GPT-4 to generate a new explanation based on the question and label without the guidance of the generic fact. Hereafter, we obtain AbsR$^*$. The only difference between AbsR and AbsR$^*$ is the explanations in AbsR are guided by the generic fact. Table~\ref{tab:ablation}~(w/ AbsR$^*$) shows the results of MeanLearn trained with AbsR$^*$, we can infer:

% To be precise, we follow the process of generating samples with GPT-4 in Sec.~\ref{sec3.1}, but we do not provide the generic fact for explanation generation. Therefore, the explanation is not guided by the generic fact. We denote this dataset as AbsR$^*$. Results are shown in Table~\ref{tab:ablation}~(w/o Guidance), from which we can infer:

(1)~Training 7B and 13B Orca-2 with AbsR$^*$ can still outperform Orca-2 on vanilla accuracy. However, it cannot bring stable improvements on \texttt{AbsAcc}. This underscores that reasoning-specific post-training can boost performance, while improving abstract reasoning performance might depend on more careful design such as MeanLearn.

(2)~MeanLearn without Knowledge outperforms MeanLearn trained with AbsR$^*$, showing the superiority of training with generic-fact-guided explanation. This stems from the generic-fact-guided explanation that can teach LLMs to use the regular pattern behind the questions for better reasoning.

\begin{table}[t]
\begin{minipage}[l]{0.48\textwidth}
    \scriptsize
    \setlength\tabcolsep{1.8pt}
    \centering
    \caption{MeanLearn trained based on Mistral.}
    \vspace{0.2cm}
    \begin{tabular}{lccccc}
    \toprule
    \textbf{Method} & \textbf{AbsR} & \textbf{Com.} & \textbf{MMLU} & \textbf{RACE} & \textit{\textbf{Average}} \\ \midrule
    \multicolumn{6}{l}{\mybox[green]{Vanilla Accuracy}} \\
    Mistral & 83.00 & 66.53 & 57.02 & 75.33 & 70.47 \\
    MeanLearn (Ours) & \textbf{84.50} & \textbf{74.56} & \textbf{58.28} & \textbf{76.65} & \textbf{73.50} \\ \midrule
    \multicolumn{6}{l}{\mybox[red]{AbsAcc}} \\
    Mistral & 72.92 & 58.87 & 40.19 & 52.69 & 56.17 \\
    MeanLearn (Ours) & \textbf{73.96} & \textbf{68.59} & \textbf{41.14} & \textbf{55.77} & \textbf{59.87} \\ \bottomrule
    \end{tabular}
    \label{tab:mistral}
\end{minipage}
\begin{minipage}[r]{0.48\textwidth}
    \scriptsize
    \setlength\tabcolsep{2.5pt}
    \centering
    \caption{Performance on mathematical reasonin datasets from MMLU~\cite{hendrycks2021measuring}.}
    \vspace{0.2cm}
    \begin{tabular}{llcccccc}
    \toprule
    \textbf{Size} & \textbf{Method} & \textbf{AA} & \textbf{CM} & \textbf{EM} & \textbf{HSS} & \textbf{HSM} & \textbf{\textit{Average}} \\ \midrule
    \multicolumn{8}{l}{\mybox[green]{Vanilla Accuracy}} \\
    \multirow{4}{*}{7B} & Orca-2 & 23.00 & 32.00 & 33.07 & 37.95 & 28.15 & 30.83 \\
     & MeanLearn & 36.00 & 35.00 & 33.07 & 40.28 & 28.89 & \textbf{34.56} \\ \cmidrule{2-8}
     & Mistral & 25.00 & 30.00 & 38.89 & 46.76 & 32.96 & 34.72 \\
     & MeanLearn & 31.00 & 34.00 & 37.83 & 47.69 & 34.07 & \textbf{36.92} \\ \midrule
    \multicolumn{8}{l}{\mybox[red]{AbsAcc}} \\
    \multirow{4}{*}{7B} & Orca-2 & 0.00 & 4.88 & 25.00 & 25.17 & 14.63 & 13.94 \\
     & MeanLearn & 2.86 & 7.32 & 24.02 & 27.97 & 16.26 & \textbf{15.69} \\ \cmidrule{2-8}
     & Mistral & 2.86 & 7.32 & 25.49 & 36.36 & 14.63 & 17.33 \\
     & MeanLearn & 5.71 & 9.76 & 24.55 & 37.76 & 16.26 & \textbf{18.81} \\ \bottomrule
    \end{tabular}
    \label{tab:math}
\end{minipage}
\end{table}

% \begin{table}[t]
% \small
% \centering
% \caption{zzz}
% \vspace{0.2cm}
% \begin{tabular}{llcccccc}
% \toprule
% \textbf{Size} & \textbf{Method} & \textbf{AA} & \textbf{CM} & \textbf{EM} & \textbf{HSS} & \textbf{HSM} & \textbf{\textit{Average}} \\ \midrule
% \multicolumn{8}{l}{\mybox[green]{Vanilla Accuracy}} \\
% \multirow{4}{*}{7B} & Orca-2 & 23.00 & 32.00 & 33.07 & 37.95 & 28.15 & 30.83 \\
%  & MeanLearn & 36.00 & 35.00 & 33.07 & 40.28 & 28.89 & 34.56 \\ \cmidrule{2-8}
%  & Mistral & 25.00 & 30.00 & 38.89 & 46.76 & 32.96 & 34.72 \\
%  & MeanLearn & 31.00 & 34.00 & 37.83 & 47.69 & 34.07 & 36.92 \\
% \multicolumn{8}{l}{\mybox[red]{AbsACC}} \\
% \multirow{4}{*}{7B} & Orca-2 & 0.00 & 4.88 & 25.00 & 25.17 & 14.63 & 13.94 \\
%  & MeanLearn & 2.86 & 7.32 & 24.02 & 27.97 & 16.26 & 15.69 \\ \cmidrule{2-8}
%  & Mistral & 2.86 & 7.32 & 25.49 & 36.36 & 14.63 & 17.33 \\
%  & MeanLearn & 5.71 & 9.76 & 24.55 & 37.76 & 16.26 & 18.81 \\ \bottomrule
% \end{tabular}
% \label{tab:my-table}
% \end{table}

\subsection{Expanding MeanLearn to LLMs besides the LLaMA series}
To investigate the universality of MeanLearn, we train MeanLearn with Mistral-7B-Instruct-v0.2~\cite{jiang2023mistral} (denoted as Mistral). All settings are consistent with Sec~\ref{sec:experiments}. We choose AbsR, Com., MMLU, and RACE for evaluation. As shown in Table~\ref{tab:mistral}, we can find MeanLearn has good applicability on Mistral.

\subsection{Expanding reasoning domains to mathematical reasoning}
Although we do not incorporate math examples in AbsR, we want to investigate whether MeanLearn can improve the performance of mathematical reasoning by enhancing abstract reasoning in LLMs. Following~\cite{luo2023wizardmath}, we select five mathematical reasoning datasets from MMLU~\cite{hendrycks2021measuring} to demonstrate this: abstract algebra~(AA), college mathematics~(CM), elementary mathematics~(EM), high school statistics~(HSS), and high school mathematics~(HSM). We choose 7B version of Orca-2 and Mistral for comparison. Results are shown in Table~\ref{tab:math}, it is interesting to find that MeanLearn can outperform baselines even if there is no math data in AbsR. This demonstrates the effectiveness of MeanLearn in enhancing abstract reasoning in LLMs.

\section{Related work}
% Our work is mainly related to the paradims of LLMs-based reasoning and distilling instruction data for post-trianing.

\subsection{Reasoning with LLMs}
LLMs have overturned the research paradigm of NLP, which makes reasoning more accessible. Using LLMs to perform reasoning can be categorized into tunning-based and prompt-based methods.

Tunning-based methods, originating from T0~\cite{victor2022multitask} and FLAN~\cite{wei2021finetuned}, finetuned T5~\cite{raffel2020exploring} and LaMDA-PT~\cite{thoppilan2022lamda} on massive NLP tasks, achieving notable zero-shot performance on unseen tasks. Flan-PaLM~\cite{chung2022scaling} and T$k$-INSTRUCT~\cite{wang2022super} further scaled up the model and task size for improvement. Orca~\cite{mukherjee2023orca} and Orca-2~\cite{mitra2023orca} leveraged vast instruction data to teach small-scale LLMs reasoning skills.
% WizardLM~\cite{xu2023wizardlm} adopted complex reasoning data to empower LLMs to follow complex instruction. Orca~\cite{mukherjee2023orca} and Orca-2~\cite{mitra2023orca} utilize numerous intruction data to teach samll-scale LLMs how to reason.

As for prompt-based methods, \citet{wei2022chain} introduced CoT prompt to elicit reasoning in LLMs by encouraging step-by-step thinking. Additionally, \citet{kojima2022large} presented zero-shot CoT to bypass manual annotation issues. Subsequently, \citet{zhang2023automatic} proposed AutoCoT to automatically generate few-shot examples with CoT for reasoning. Various CoT-based methods~\cite{jin2024self,wang2024strategic} have since emerged, such as complex CoT~\cite{fu2022complexity}, and the tree of thought~\cite{yao2023tree}, etc.  
% Furthermore, some LLMs collaboration methods are proposed for reasoning~\cite{xiong2023examining,du2023improving}.

These methods are designed for improving the general reasoning abilities of LLMs, while we focus on the abstract reasoning capabilities of LLMs.

\subsection{LLMs-as-annotators}
LLMs are increasingly used as annotators to create data for training and evaluation due to their advanced instruction-following skills.
% LLMs are increasingly employed for data labelling due to their advanced ability to follow instructions.

Some studies distilled various CoT data from LLMs to train student models~(\cite{li-etal-2023-symbolic,ho-etal-2023-large,ying2024llms}, \textit{inter alia}), achieving impressive performance. For instance, \citet{li-etal-2023-symbolic} distilled symbolic CoT from GPT-3 and observed enhanced performance in commonsense reasoning. \citet{dai2023auggpt} used ChatGPT to generate augmentation data to enhance BERT~\cite{kenton2019bert}. With the advent of open-source LLMs like LLaMA~\cite{touvron2023llama}, other studies distilled instruction data from large-scale LLMs to train small-scale models~(\cite{xu2023wizardlm,mukherjee2023orca,kim2024small,liao2024neural,just2024dipt,gupta2024selective}, \textit{inter alia}). For example,  \citet{xu2023wizardlm} and \citet{luo2023wizardmath} employed an evolving strategy to obtain complex data, trained LLaMA, and achieved strong performance. \citet{tangmathscale} constructed a concept graph and used GPT-3.5 to synthesize math data with high diversity. We employ LLMs to generate training data and develop AbsR tailored for abstract reasoning.

% For evaluation purposes, some works asked LLMs to annotate data to test specific or general abilities of existing models~(\citeauthor{bai2023benchmarking}, \citeyear{bai2023benchmarking}; \citeauthor{ying2023intuitive}, \citeyear{ying2023intuitive}; \textit{inter alia}). For example, \citet{ying2023intuitive} utilize LLMs to construct multipul-choice questions for more precise evaluation.

% Our work is similar to adopting LLMs to generate training data, and we create AbsR tailored for abstract reasoning.
% We employ LLMs to generate training data and develop AbsR tailored for abstract reasoning.

\section{Conclusion}
In this paper, we investigate the abstract reasoning of existing LLMs. To achieve this, we first define the evaluation metric of abstract reasoning and design a series of preliminary experiments. Then we discover a significant performance gap between general reasoning and abstract reasoning. Hence, we tailor an abstract reasoning dataset AbsR with the help of GPT-4 to enhance LLMs. Finally, we devise a simple but effective paradigm~(MeanLearn) to teach LLMs abstract reasoning in \textit{\textbf{Knowledge}} and \textit{\textbf{Reasoning}} ways. Extensive experiments demonstrate the superiority and effectiveness of MeanLearn in vanilla and abstract reasoning accuracies. 
The limitations of MeanLearn are discussed in Appendix~\ref{app:limitations}.

\section*{Acknowledgements}
We would like to acknowledge the editors and reviewers for their efforts and advice, and gratefully acknowledge the support of the National Natural Science Foundation of China under Grants U22B2059 and 62176079, Natural Science Foundation of Heilongjiang Province under Grant YQ2022F005.

\bibliography{neurips_2024}

\begin{thebibliography}{60}
\providecommand{\natexlab}[1]{#1}
\providecommand{\url}[1]{\texttt{#1}}
\expandafter\ifx\csname urlstyle\endcsname\relax
  \providecommand{\doi}[1]{doi: #1}\else
  \providecommand{\doi}{doi: \begingroup \urlstyle{rm}\Url}\fi

\bibitem[sha(2023)]{sharegpt}
Sharegpt: Share your wildest chatgpt conversations with one click, 2023.

\bibitem[Achiam et~al.(2023)Achiam, Adler, Agarwal, Ahmad, Akkaya, Aleman, Almeida, Altenschmidt, Altman, Anadkat, et~al.]{achiam2023gpt}
Josh Achiam, Steven Adler, Sandhini Agarwal, Lama Ahmad, Ilge Akkaya, Florencia~Leoni Aleman, Diogo Almeida, Janko Altenschmidt, Sam Altman, Shyamal Anadkat, et~al.
\newblock Gpt-4 technical report.
\newblock \emph{arXiv preprint arXiv:2303.08774}, 2023.

\bibitem[Amayuelas et~al.(2023)Amayuelas, Pan, Chen, and Wang]{amayuelas2023knowledge}
Alfonso Amayuelas, Liangming Pan, Wenhu Chen, and William Wang.
\newblock Knowledge of knowledge: Exploring known-unknowns uncertainty with large language models.
\newblock \emph{arXiv preprint arXiv:2305.13712}, 2023.

\bibitem[Bhagavatula et~al.(2019)Bhagavatula, Le~Bras, Malaviya, Sakaguchi, Holtzman, Rashkin, Downey, Yih, and Choi]{bhagavatula2019abductive}
Chandra Bhagavatula, Ronan Le~Bras, Chaitanya Malaviya, Keisuke Sakaguchi, Ari Holtzman, Hannah Rashkin, Doug Downey, Wen-tau Yih, and Yejin Choi.
\newblock Abductive commonsense reasoning.
\newblock In \emph{International Conference on Learning Representations}, 2019.

\bibitem[Bhakthavatsalam et~al.(2020)Bhakthavatsalam, Anastasiades, and Clark]{bhakthavatsalam2020genericskb}
Sumithra Bhakthavatsalam, Chloe Anastasiades, and Peter Clark.
\newblock Genericskb: A knowledge base of generic statements.
\newblock \emph{arXiv preprint arXiv:2005.00660}, 2020.

\bibitem[Bisk et~al.(2020)Bisk, Zellers, Gao, Choi, et~al.]{bisk2020piqa}
Yonatan Bisk, Rowan Zellers, Jianfeng Gao, Yejin Choi, et~al.
\newblock Piqa: Reasoning about physical commonsense in natural language.
\newblock In \emph{Proceedings of the AAAI conference on artificial intelligence}, volume~34, pages 7432--7439, 2020.

\bibitem[Brown et~al.(2020)Brown, Mann, Ryder, Subbiah, Kaplan, Dhariwal, Neelakantan, Shyam, Sastry, Askell, et~al.]{brown2020language}
Tom Brown, Benjamin Mann, Nick Ryder, Melanie Subbiah, Jared~D Kaplan, Prafulla Dhariwal, Arvind Neelakantan, Pranav Shyam, Girish Sastry, Amanda Askell, et~al.
\newblock Language models are few-shot learners.
\newblock \emph{Advances in neural information processing systems}, 33:\penalty0 1877--1901, 2020.

\bibitem[Chiang et~al.(2023)Chiang, Li, Lin, Sheng, Wu, Zhang, Zheng, Zhuang, Zhuang, Gonzalez, Stoica, and Xing]{vicuna2023}
Wei-Lin Chiang, Zhuohan Li, Zi~Lin, Ying Sheng, Zhanghao Wu, Hao Zhang, Lianmin Zheng, Siyuan Zhuang, Yonghao Zhuang, Joseph~E. Gonzalez, Ion Stoica, and Eric~P. Xing.
\newblock Vicuna: An open-source chatbot impressing gpt-4 with 90\%* chatgpt quality, March 2023.

\bibitem[Chung et~al.(2022)Chung, Hou, Longpre, Zoph, Tay, Fedus, Li, Wang, Dehghani, Brahma, et~al.]{chung2022scaling}
Hyung~Won Chung, Le~Hou, Shayne Longpre, Barret Zoph, Yi~Tay, William Fedus, Yunxuan Li, Xuezhi Wang, Mostafa Dehghani, Siddhartha Brahma, et~al.
\newblock Scaling instruction-finetuned language models.
\newblock \emph{arXiv preprint arXiv:2210.11416}, 2022.

\bibitem[Clark et~al.(2018)Clark, Cowhey, Etzioni, Khot, Sabharwal, Schoenick, and Tafjord]{clark2018think}
Peter Clark, Isaac Cowhey, Oren Etzioni, Tushar Khot, Ashish Sabharwal, Carissa Schoenick, and Oyvind Tafjord.
\newblock Think you have solved question answering? try arc, the ai2 reasoning challenge.
\newblock \emph{arXiv preprint arXiv:1803.05457}, 2018.

\bibitem[Contributors(2023)]{2023opencompass}
OpenCompass Contributors.
\newblock Opencompass: A universal evaluation platform for foundation models, 2023.

\bibitem[Dai et~al.(2023)Dai, Liu, Liao, Huang, Cao, Wu, Zhao, Xu, Liu, Liu, et~al.]{dai2023auggpt}
Haixing Dai, Zhengliang Liu, Wenxiong Liao, Xiaoke Huang, Yihan Cao, Zihao Wu, Lin Zhao, Shaochen Xu, Wei Liu, Ninghao Liu, et~al.
\newblock Auggpt: Leveraging chatgpt for text data augmentation.
\newblock \emph{arXiv preprint arXiv:2302.13007}, 2023.

\bibitem[Du et~al.(2019)Du, Ding, Liu, and Li]{du2019modeling}
Li~Du, Xiao Ding, Ting Liu, and Zhongyang Li.
\newblock Modeling event background for if-then commonsense reasoning using context-aware variational autoencoder.
\newblock In \emph{Proceedings of the 2019 Conference on Empirical Methods in Natural Language Processing and the 9th International Joint Conference on Natural Language Processing (EMNLP-IJCNLP)}, pages 2682--2691, 2019.

\bibitem[Du et~al.(2022)Du, Ding, Xiong, Liu, and Qin]{du2022care}
Li~Du, Xiao Ding, Kai Xiong, Ting Liu, and Bing Qin.
\newblock e-care: a new dataset for exploring explainable causal reasoning.
\newblock In \emph{Proceedings of the 60th Annual Meeting of the Association for Computational Linguistics (Volume 1: Long Papers)}, pages 432--446, 2022.

\bibitem[Fu et~al.(2023)Fu, Peng, Sabharwal, Clark, and Khot]{fu2022complexity}
Yao Fu, Hao Peng, Ashish Sabharwal, Peter Clark, and Tushar Khot.
\newblock Complexity-based prompting for multi-step reasoning.
\newblock In \emph{The Eleventh International Conference on Learning Representations}, 2023.

\bibitem[Geva et~al.(2021)Geva, Khashabi, Segal, Khot, Roth, and Berant]{geva2021did}
Mor Geva, Daniel Khashabi, Elad Segal, Tushar Khot, Dan Roth, and Jonathan Berant.
\newblock Did aristotle use a laptop? a question answering benchmark with implicit reasoning strategies.
\newblock \emph{Transactions of the Association for Computational Linguistics}, 9:\penalty0 346--361, 2021.

\bibitem[Gupta et~al.(2024)Gupta, Nandwani, Yehudai, Mishra, Pandey, Raghu, and Joshi]{gupta2024selective}
Sonam Gupta, Yatin Nandwani, Asaf Yehudai, Mayank Mishra, Gaurav Pandey, Dinesh Raghu, and Sachindra Joshi.
\newblock Selective self-rehearsal: A fine-tuning approach to improve generalization in large language models.
\newblock \emph{arXiv preprint arXiv:2409.04787}, 2024.

\bibitem[Hendrycks et~al.(2021)Hendrycks, Burns, Basart, Zou, Mazeika, Song, and Steinhardt]{hendrycks2021measuring}
Dan Hendrycks, Collin Burns, Steven Basart, Andy Zou, Mantas Mazeika, Dawn Song, and Jacob Steinhardt.
\newblock Measuring massive multitask language understanding.
\newblock In \emph{International Conference on Learning Representations}, 2021.

\bibitem[Ho et~al.(2023)Ho, Schmid, and Yun]{ho-etal-2023-large}
Namgyu Ho, Laura Schmid, and Se-Young Yun.
\newblock Large language models are reasoning teachers.
\newblock In \emph{Proceedings of the 61st Annual Meeting of the Association for Computational Linguistics (Volume 1: Long Papers)}, pages 14852--14882, Toronto, Canada, July 2023. Association for Computational Linguistics.

\bibitem[Hu et~al.(2022)Hu, Wallis, Allen-Zhu, Li, Wang, Wang, Chen, et~al.]{hu2022lora}
Edward~J Hu, Phillip Wallis, Zeyuan Allen-Zhu, Yuanzhi Li, Shean Wang, Lu~Wang, Weizhu Chen, et~al.
\newblock Lora: Low-rank adaptation of large language models.
\newblock In \emph{International Conference on Learning Representations}, 2022.

\bibitem[Jiang et~al.(2023)Jiang, Sablayrolles, Mensch, Bamford, Chaplot, Casas, Bressand, Lengyel, Lample, Saulnier, et~al.]{jiang2023mistral}
Albert~Q Jiang, Alexandre Sablayrolles, Arthur Mensch, Chris Bamford, Devendra~Singh Chaplot, Diego de~las Casas, Florian Bressand, Gianna Lengyel, Guillaume Lample, Lucile Saulnier, et~al.
\newblock Mistral 7b.
\newblock \emph{arXiv preprint arXiv:2310.06825}, 2023.

\bibitem[Jin and Lu(2024)]{jin2024self}
Ziqi Jin and Wei Lu.
\newblock Self-harmonized chain of thought.
\newblock \emph{arXiv preprint arXiv:2409.04057}, 2024.

\bibitem[Just et~al.(2024)Just, Dabas, Huang, Jin, and Jia]{just2024dipt}
Hoang~Anh Just, Mahavir Dabas, Lifu Huang, Ming Jin, and Ruoxi Jia.
\newblock Dipt: Enhancing llm reasoning through diversified perspective-taking.
\newblock \emph{arXiv preprint arXiv:2409.06241}, 2024.

\bibitem[Kenton and Toutanova(2019)]{kenton2019bert}
Jacob Devlin Ming-Wei~Chang Kenton and Lee~Kristina Toutanova.
\newblock Bert: Pre-training of deep bidirectional transformers for language understanding.
\newblock In \emph{Proceedings of naacL-HLT}, volume~1, page~2. Minneapolis, Minnesota, 2019.

\bibitem[Kim et~al.(2024)Kim, Lee, Kim, and Lee]{kim2024small}
Bumjun Kim, Kunha Lee, Juyeon Kim, and Sangam Lee.
\newblock Small language models are equation reasoners.
\newblock \emph{arXiv preprint arXiv:2409.12393}, 2024.

\bibitem[Kingma and Ba(2014)]{kingma2014adam}
Diederik~P Kingma and Jimmy Ba.
\newblock Adam: A method for stochastic optimization.
\newblock \emph{arXiv preprint arXiv:1412.6980}, 2014.

\bibitem[Kojima et~al.(2022)Kojima, Gu, Reid, Matsuo, and Iwasawa]{kojima2022large}
Takeshi Kojima, Shixiang~Shane Gu, Machel Reid, Yutaka Matsuo, and Yusuke Iwasawa.
\newblock Large language models are zero-shot reasoners.
\newblock \emph{Advances in neural information processing systems}, 35:\penalty0 22199--22213, 2022.

\bibitem[Lai et~al.(2017)Lai, Xie, Liu, Yang, and Hovy]{lai2017race}
Guokun Lai, Qizhe Xie, Hanxiao Liu, Yiming Yang, and Eduard Hovy.
\newblock Race: Large-scale reading comprehension dataset from examinations.
\newblock In \emph{Proceedings of the 2017 Conference on Empirical Methods in Natural Language Processing}, pages 785--794, 2017.

\bibitem[Lake et~al.(2017)Lake, Ullman, Tenenbaum, and Gershman]{lake2017building}
Brenden~M Lake, Tomer~D Ullman, Joshua~B Tenenbaum, and Samuel~J Gershman.
\newblock Building machines that learn and think like people.
\newblock \emph{Behavioral and brain sciences}, 40:\penalty0 e253, 2017.

\bibitem[Li et~al.(2023)Li, Hessel, Yu, Ren, Chang, and Choi]{li-etal-2023-symbolic}
Liunian~Harold Li, Jack Hessel, Youngjae Yu, Xiang Ren, Kai-Wei Chang, and Yejin Choi.
\newblock Symbolic chain-of-thought distillation: Small models can also {``}think{''} step-by-step.
\newblock In \emph{Proceedings of the 61st Annual Meeting of the Association for Computational Linguistics (Volume 1: Long Papers)}, pages 2665--2679, Toronto, Canada, July 2023. Association for Computational Linguistics.

\bibitem[Liao et~al.(2024)Liao, He, Xu, Zhang, Liu, and Zhao]{liao2024neural}
Huanxuan Liao, Shizhu He, Yao Xu, Yuanzhe Zhang, Kang Liu, and Jun Zhao.
\newblock Neural-symbolic collaborative distillation: Advancing small language models for complex reasoning tasks.
\newblock \emph{arXiv preprint arXiv:2409.13203}, 2024.

\bibitem[Liu et~al.(2019)Liu, Ott, Goyal, Du, Joshi, Chen, Levy, Lewis, Zettlemoyer, and Stoyanov]{liu2019roberta}
Yinhan Liu, Myle Ott, Naman Goyal, Jingfei Du, Mandar Joshi, Danqi Chen, Omer Levy, Mike Lewis, Luke Zettlemoyer, and Veselin Stoyanov.
\newblock Roberta: A robustly optimized bert pretraining approach.
\newblock \emph{arXiv preprint arXiv:1907.11692}, 2019.

\bibitem[Luo et~al.(2023)Luo, Sun, Xu, Zhao, Lou, Tao, Geng, Lin, Chen, and Zhang]{luo2023wizardmath}
Haipeng Luo, Qingfeng Sun, Can Xu, Pu~Zhao, Jianguang Lou, Chongyang Tao, Xiubo Geng, Qingwei Lin, Shifeng Chen, and Dongmei Zhang.
\newblock Wizardmath: Empowering mathematical reasoning for large language models via reinforced evol-instruct.
\newblock \emph{arXiv preprint arXiv:2308.09583}, 2023.

\bibitem[{Meta AI}(2024)]{meta2024introducing}
{Meta AI}.
\newblock Introducing meta llama 3: The most capable openly available llm to date, 2024.

\bibitem[Mitra et~al.(2023)Mitra, Del~Corro, Mahajan, Codas, Simoes, Agarwal, Chen, Razdaibiedina, Jones, Aggarwal, et~al.]{mitra2023orca}
Arindam Mitra, Luciano Del~Corro, Shweti Mahajan, Andres Codas, Clarisse Simoes, Sahaj Agarwal, Xuxi Chen, Anastasia Razdaibiedina, Erik Jones, Kriti Aggarwal, et~al.
\newblock Orca 2: Teaching small language models how to reason.
\newblock \emph{arXiv preprint arXiv:2311.11045}, 2023.

\bibitem[Mukherjee et~al.(2023)Mukherjee, Mitra, Jawahar, Agarwal, Palangi, and Awadallah]{mukherjee2023orca}
Subhabrata Mukherjee, Arindam Mitra, Ganesh Jawahar, Sahaj Agarwal, Hamid Palangi, and Ahmed Awadallah.
\newblock Orca: Progressive learning from complex explanation traces of gpt-4.
\newblock \emph{arXiv preprint arXiv:2306.02707}, 2023.

\bibitem[Qiu et~al.(2023)Qiu, Jiang, Lu, Sclar, Pyatkin, Bhagavatula, Wang, Kim, Choi, Dziri, et~al.]{qiu2023phenomenal}
Linlu Qiu, Liwei Jiang, Ximing Lu, Melanie Sclar, Valentina Pyatkin, Chandra Bhagavatula, Bailin Wang, Yoon Kim, Yejin Choi, Nouha Dziri, et~al.
\newblock Phenomenal yet puzzling: Testing inductive reasoning capabilities of language models with hypothesis refinement.
\newblock \emph{arXiv preprint arXiv:2310.08559}, 2023.

\bibitem[Raffel et~al.(2020)Raffel, Shazeer, Roberts, Lee, Narang, Matena, Zhou, Li, and Liu]{raffel2020exploring}
Colin Raffel, Noam Shazeer, Adam Roberts, Katherine Lee, Sharan Narang, Michael Matena, Yanqi Zhou, Wei Li, and Peter~J Liu.
\newblock Exploring the limits of transfer learning with a unified text-to-text transformer.
\newblock \emph{The Journal of Machine Learning Research}, 21\penalty0 (1):\penalty0 5485--5551, 2020.

\bibitem[Roemmele et~al.(2011)Roemmele, Bejan, and Gordon]{roemmele2011choice}
Melissa Roemmele, Cosmin~Adrian Bejan, and Andrew~S Gordon.
\newblock Choice of plausible alternatives: An evaluation of commonsense causal reasoning.
\newblock In \emph{2011 AAAI Spring Symposium Series}, 2011.

\bibitem[Sap et~al.(2019)Sap, Rashkin, Chen, Le~Bras, and Choi]{sap2019social}
Maarten Sap, Hannah Rashkin, Derek Chen, Ronan Le~Bras, and Yejin Choi.
\newblock Social iqa: Commonsense reasoning about social interactions.
\newblock In \emph{Proceedings of the 2019 Conference on Empirical Methods in Natural Language Processing and the 9th International Joint Conference on Natural Language Processing (EMNLP-IJCNLP)}, pages 4463--4473, 2019.

\bibitem[Shi et~al.(2021)Shi, Ding, Du, Liu, and Qin]{shi2021neural}
Jihao Shi, Xiao Ding, Li~Du, Ting Liu, and Bing Qin.
\newblock Neural natural logic inference for interpretable question answering.
\newblock In \emph{Proceedings of the 2021 Conference on Empirical Methods in Natural Language Processing}, pages 3673--3684, 2021.

\bibitem[Srivastava et~al.(2023)Srivastava, Rastogi, Rao, Shoeb, Abid, Fisch, Brown, Santoro, Gupta, Garriga-Alonso, et~al.]{srivastava2023beyond}
Aarohi Srivastava, Abhinav Rastogi, Abhishek Rao, Abu Awal~Md Shoeb, Abubakar Abid, Adam Fisch, Adam~R Brown, Adam Santoro, Aditya Gupta, Adri{\`a} Garriga-Alonso, et~al.
\newblock Beyond the imitation game: Quantifying and extrapolating the capabilities of language models.
\newblock \emph{Transactions on Machine Learning Research}, 2023.

\bibitem[Suzgun et~al.(2023)Suzgun, Scales, Sch{\"a}rli, Gehrmann, Tay, Chung, Chowdhery, Le, Chi, Zhou, and Wei]{suzgun-etal-2023-challenging}
Mirac Suzgun, Nathan Scales, Nathanael Sch{\"a}rli, Sebastian Gehrmann, Yi~Tay, Hyung~Won Chung, Aakanksha Chowdhery, Quoc Le, Ed~Chi, Denny Zhou, and Jason Wei.
\newblock Challenging {BIG}-bench tasks and whether chain-of-thought can solve them.
\newblock In Anna Rogers, Jordan Boyd-Graber, and Naoaki Okazaki, editors, \emph{Findings of the Association for Computational Linguistics: ACL 2023}, pages 13003--13051, Toronto, Canada, July 2023. Association for Computational Linguistics.

\bibitem[Talmor et~al.(2019)Talmor, Herzig, Lourie, and Berant]{talmor2019commonsenseqa}
Alon Talmor, Jonathan Herzig, Nicholas Lourie, and Jonathan Berant.
\newblock Commonsenseqa: A question answering challenge targeting commonsense knowledge.
\newblock In \emph{Proceedings of the 2019 Conference of the North American Chapter of the Association for Computational Linguistics: Human Language Technologies, Volume 1 (Long and Short Papers)}, pages 4149--4158, 2019.

\bibitem[Tang et~al.(2024)Tang, Zhang, Wang, and Wei]{tangmathscale}
Zhengyang Tang, Xingxing Zhang, Benyou Wang, and Furu Wei.
\newblock Mathscale: Scaling instruction tuning for mathematical reasoning.
\newblock In \emph{Forty-first International Conference on Machine Learning}, 2024.

\bibitem[Thoppilan et~al.(2022)Thoppilan, De~Freitas, Hall, Shazeer, Kulshreshtha, Cheng, Jin, Bos, Baker, Du, et~al.]{thoppilan2022lamda}
Romal Thoppilan, Daniel De~Freitas, Jamie Hall, Noam Shazeer, Apoorv Kulshreshtha, Heng-Tze Cheng, Alicia Jin, Taylor Bos, Leslie Baker, Yu~Du, et~al.
\newblock Lamda: Language models for dialog applications.
\newblock \emph{arXiv preprint arXiv:2201.08239}, 2022.

\bibitem[Touvron et~al.(2023)Touvron, Martin, Stone, Albert, Almahairi, Babaei, Bashlykov, Batra, Bhargava, Bhosale, et~al.]{touvron2023llama}
Hugo Touvron, Louis Martin, Kevin Stone, Peter Albert, Amjad Almahairi, Yasmine Babaei, Nikolay Bashlykov, Soumya Batra, Prajjwal Bhargava, Shruti Bhosale, et~al.
\newblock Llama 2: Open foundation and fine-tuned chat models.
\newblock \emph{arXiv preprint arXiv:2307.09288}, 2023.

\bibitem[Victor et~al.(2022)Victor, Albert, Colin, Stephen, Lintang, Zaid, Antoine, Arnaud, Arun, Manan, et~al.]{victor2022multitask}
Sanh Victor, Webson Albert, Raffel Colin, Bach Stephen, Sutawika Lintang, Alyafeai Zaid, Chaffin Antoine, Stiegler Arnaud, Raja Arun, Dey Manan, et~al.
\newblock Multitask prompted training enables zero-shot task generalization.
\newblock In \emph{International Conference on Learning Representations}, 2022.

\bibitem[Wang et~al.(2022{\natexlab{a}})Wang, Kordi, Mishra, Liu, Smith, Khashabi, and Hajishirzi]{wang2022self}
Yizhong Wang, Yeganeh Kordi, Swaroop Mishra, Alisa Liu, Noah~A Smith, Daniel Khashabi, and Hannaneh Hajishirzi.
\newblock Self-instruct: Aligning language model with self generated instructions.
\newblock \emph{arXiv preprint arXiv:2212.10560}, 2022{\natexlab{a}}.

\bibitem[Wang et~al.(2022{\natexlab{b}})Wang, Mishra, Alipoormolabashi, Kordi, Mirzaei, Naik, Ashok, Dhanasekaran, Arunkumar, Stap, et~al.]{wang2022super}
Yizhong Wang, Swaroop Mishra, Pegah Alipoormolabashi, Yeganeh Kordi, Amirreza Mirzaei, Atharva Naik, Arjun Ashok, Arut~Selvan Dhanasekaran, Anjana Arunkumar, David Stap, et~al.
\newblock Super-naturalinstructions: Generalization via declarative instructions on 1600+ nlp tasks.
\newblock In \emph{Proceedings of the 2022 Conference on Empirical Methods in Natural Language Processing}, pages 5085--5109, 2022{\natexlab{b}}.

\bibitem[Wang et~al.(2024)Wang, Zhao, Wang, Huang, Fan, Zhang, Wang, Wang, and Liu]{wang2024strategic}
Yu~Wang, Shiwan Zhao, Zhihu Wang, Heyuan Huang, Ming Fan, Yubo Zhang, Zhixing Wang, Haijun Wang, and Ting Liu.
\newblock Strategic chain-of-thought: Guiding accurate reasoning in llms through strategy elicitation.
\newblock \emph{arXiv preprint arXiv:2409.03271}, 2024.

\bibitem[Wei et~al.(2022{\natexlab{a}})Wei, Bosma, Zhao, Guu, Yu, Lester, Du, Dai, and Le]{wei2021finetuned}
Jason Wei, Maarten Bosma, Vincent Zhao, Kelvin Guu, Adams~Wei Yu, Brian Lester, Nan Du, Andrew~M Dai, and Quoc~V Le.
\newblock Finetuned language models are zero-shot learners.
\newblock In \emph{International Conference on Learning Representations}, 2022{\natexlab{a}}.

\bibitem[Wei et~al.(2022{\natexlab{b}})Wei, Wang, Schuurmans, Bosma, Xia, Chi, Le, Zhou, et~al.]{wei2022chain}
Jason Wei, Xuezhi Wang, Dale Schuurmans, Maarten Bosma, Fei Xia, Ed~Chi, Quoc~V Le, Denny Zhou, et~al.
\newblock Chain-of-thought prompting elicits reasoning in large language models.
\newblock \emph{Advances in Neural Information Processing Systems}, 35:\penalty0 24824--24837, 2022{\natexlab{b}}.

\bibitem[Xiong et~al.(2022)Xiong, Ding, Li, Du, Liu, Qin, Zheng, and Huai]{xiong2022reco}
Kai Xiong, Xiao Ding, Zhongyang Li, Li~Du, Ting Liu, Bing Qin, Yi~Zheng, and Baoxing Huai.
\newblock Reco: Reliable causal chain reasoning via structural causal recurrent neural networks.
\newblock In \emph{Proceedings of the 2022 Conference on Empirical Methods in Natural Language Processing}, pages 6426--6438, 2022.

\bibitem[Xiong et~al.(2023)Xiong, Ding, Cao, Liu, and Qin]{xiong2023examining}
Kai Xiong, Xiao Ding, Yixin Cao, Ting Liu, and Bing Qin.
\newblock Examining inter-consistency of large language models collaboration: An in-depth analysis via debate.
\newblock In \emph{Findings of the Association for Computational Linguistics: EMNLP 2023}, pages 7572--7590, 2023.

\bibitem[Xu et~al.(2023)Xu, Sun, Zheng, Geng, Zhao, Feng, Tao, and Jiang]{xu2023wizardlm}
Can Xu, Qingfeng Sun, Kai Zheng, Xiubo Geng, Pu~Zhao, Jiazhan Feng, Chongyang Tao, and Daxin Jiang.
\newblock Wizardlm: Empowering large language models to follow complex instructions.
\newblock \emph{arXiv preprint arXiv:2304.12244}, 2023.

\bibitem[Yao et~al.(2023)Yao, Yu, Zhao, Shafran, Griffiths, Cao, and Narasimhan]{yao2023tree}
Shunyu Yao, Dian Yu, Jeffrey Zhao, Izhak Shafran, Thomas~L Griffiths, Yuan Cao, and Karthik Narasimhan.
\newblock Tree of thoughts: Deliberate problem solving with large language models.
\newblock \emph{arXiv preprint arXiv:2305.10601}, 2023.

\bibitem[Ying et~al.(2024)Ying, Lin, Cao, Tang, Wang, Sun, Huang, and Yan]{ying2024llms}
Jiahao Ying, Mingbao Lin, Yixin Cao, Wei Tang, Bo~Wang, Qianru Sun, Xuanjing Huang, and Shuicheng Yan.
\newblock Llms-as-instructors: Learning from errors toward automating model improvement.
\newblock \emph{Findings of EMNLP}, 2024.

\bibitem[Zhang et~al.(2023)Zhang, Zhang, Li, and Smola]{zhang2023automatic}
Zhuosheng Zhang, Aston Zhang, Mu~Li, and Alex Smola.
\newblock Automatic chain of thought prompting in large language models.
\newblock In \emph{The Eleventh International Conference on Learning Representations}, 2023.

\bibitem[Zhong et~al.(2023)Zhong, Cui, Guo, Liang, Lu, Wang, Saied, Chen, and Duan]{zhong2023agieval}
Wanjun Zhong, Ruixiang Cui, Yiduo Guo, Yaobo Liang, Shuai Lu, Yanlin Wang, Amin Saied, Weizhu Chen, and Nan Duan.
\newblock Agieval: A human-centric benchmark for evaluating foundation models.
\newblock \emph{arXiv preprint arXiv:2304.06364}, 2023.

\end{thebibliography}
\bibliographystyle{plainnat}

\newpage
\appendix
\section{Limitations}
\label{app:limitations}
There are several limitations of our work: First, the size of the training data should be larger to obtain better and more stable results. Second, there should be some examples~(whether MeanLearn can generate meaningful explanations) to demonstrate the superiority of MeanLearn. Third, more efforts should be made to focus more on abstract reasoning itself. Finally, complex scenarios~(with complex generic facts) such as causal chain reasoning~\cite{xiong2022reco} and logical reasoning~\cite{shi2021neural} is a potential direction of abstract reasoning.

\section{Asbstract reasoning study}
\subsection{Prompt for experiment I}
\label{app:abs_rea}
\subsubsection{Orca-2}
\begin{tcolorbox}[fontupper = \ttfamily]
    <|im\_start|>\mybox[green]{system}\\
    You are Orca, an AI language model created by Microsoft. You are a cautious assistant. You carefully follow instructions. You are helpful and harmless and you follow ethical guidelines and promote positive behavior.<|im\_end|>\\
    <|im\_start|>\mybox[yellow]{user}\\
    \mybox[yellow]{\{premise\}} Hypothesis1 or Hypothesis2?\\Hypothesis1: \mybox[yellow]{\{hypothesis1\}}\\Hypothesis2: \mybox[yellow]{\{hypothesis2\}} \\Your answer should follow the format like ``Answer: Hypothesis(1 or 2) is more plausible.\\Explanation: \_\_\_''<|im\_end|>\\
    <|im\_start|>\mybox[red]{assistant}
\end{tcolorbox}

\subsubsection{LLaMA-2 and GPT-3.5}
\begin{tcolorbox}[fontupper = \ttfamily]
     \mybox[green]{System}: You are a helpful assistant.\\
     \mybox[yellow]{User}: \mybox[yellow]{\{premise\}} Hypothesis1 or Hypothesis2?\\Hypothesis1: \mybox[yellow]{\{hypothesis1\}}\\Hypothesis2: \mybox[yellow]{\{hypothesis2\}} \\Your answer should follow the format like ``Answer: Hypothesis(1 or 2) is more plausible.\\Explanation: \_\_\_''
\end{tcolorbox}

The placeholders \mybox[yellow]{\{premise\}}, \mybox[yellow]{\{hypothesis1\}}, and \mybox[yellow]{\{hypothesis2\}} will be filled with the corresponding terms in each example of e-CARE dataset.

\subsection{Prompt for experiment II: Orca-2}
\label{app:fact_probing}
\subsubsection{Orca-2}
\begin{tcolorbox}[fontupper = \ttfamily]
    <|im\_start|>\mybox[green]{system}\\
    You are Orca, an AI language model created by Microsoft. You are a cautious assistant. You carefully follow instructions. You are helpful and harmless and you follow ethical guidelines and promote positive behavior.<|im\_end|>\\
    <|im\_start|>\mybox[yellow]{user}\\
    You are given a fact, do you know this fact? Just answer Yes or No, do not give any additional information.\\
    Fact: \mybox[yellow]{\{fact\}} <|im\_end|>\\
    <|im\_start|>\mybox[red]{assistant}
\end{tcolorbox}

\subsubsection{LLaMA-2 and GPT-3.5}
\begin{tcolorbox}[fontupper = \ttfamily]
     \mybox[green]{System}: You are a helpful assistant.\\
     \mybox[yellow]{User}: You are given a fact, do you know this fact? Just answer Yes or No, do not give any additional information.\\
     Fact:  \mybox[yellow]{\{fact\}} 
\end{tcolorbox}

The placeholders \mybox[yellow]{\{premise\}}, \mybox[yellow]{\{hypothesis1\}}, \mybox[yellow]{\{hypothesis1\}}, and \mybox[yellow]{\{fact\}} will be filled with the corresponding terms in each example of e-CARE dataset.

\section{Details for generic facts filtering}
\label{details:fact_filtering}
We apply the following steps to filter generics facts from GenericsKB:

(1)~Exclude sentences with confidence $<$ 0.7.

(2)~Categorize the sentences by the concepts.

(3)~Randomly sample 4,613 concepts, and then randomly sample one sentence as the generic fact from the category of each sampled concept.

Hereafter, we can obtain a collection of 4,613 generic facts.

\section{Prompt for AbsR construction}
\label{pmp:dataset_construction}
We use the following prompt to generate our AbsR examples.
\begin{tcolorbox}[]
    You are an expert in creating questions, you should first offer a question together with some options based on the fact the user gives. Second, you should give an answer and an explanation guided by the given fact. You can propose questions in any area, including but not limited to history, law, medicine, math, science, computer science, psychology, AI, politics, economics, etc. Your response should follow the following format: ``Question: \_\_\_\_
Options:\_\_\_\_\_
Answer: \_\_\_\_\_
Explanation: \_\_\_\_\_''. NOTE that the fact should be an implicit explanation for obtaining the true answer, which means the fact SHOULD NOT appear explicitly in the questions or the options. The explanations should be short. Please create \mybox[red]{\{number\}} examples.
\end{tcolorbox}
The placeholder \mybox[red]{\{number\}} will be randomly filled with ``one'', ``two'' or ``three''.

\section{Details of human evaluation}
\label{app:human}
We choose three annotators with good backgrounds in textual inference and event reasoning. The whole test set of AbsR is involved in the human evaluation (200 instances). For each annotator, we pay \$10 per hour, while in our country, the minimum wage is less than \$5 per hour. All annotators agreed to let us use their annotations. The evaluation agreements are 95\%, 91\%, and 90\% for the question answering, generic fact supportance, and diversity tasks, respectively.

\section{Instructions of human evaluation}
\label{app:instruction}
\subsection{Performance}
You are given some questions each with some options, all the questions are about commonsense and are generated by GPT-4. Please choose the most plausible option for each question. Just type your choice~(such as (A)) in the Answer column.

\subsection{Support rate}
You are given some questions each with some options and a generic fact, all the questions are about commonsense and are generated by GPT-4. Please judge whether the generic can support to answer the question. Just type your choice~(Yes or No) in the Answer column.

\subsection{Diversity}
You are given some generic facts, each generic fact possesses several samples~(each sample consists of a question and a few options). The samples can be supported by the corresponding generic fact. Please judge whether the samples within a generic are similar and are just different in nouns or expressions. If sample 1 and sample 2 are similar and are just different in nouns or expressions, just type a tuple (1, 2) in the Answer column. If there are multiple similar sample pairs, just separate them with a ``$\backslash$n''. For example, (1, 2)$\backslash$n(3, 4).

\section{K-example and R-example}
\label{exp:examples}
\subsection{K-example}
\begin{tcolorbox}[fontupper = \ttfamily]
    <|im\_start|>\mybox[green]{system}\\
    You are a cautious assistant. You carefully follow instructions. You are given a question, a few options, and a rule, you should follow the given rule to answer the question. Your response should follow the format like ``Explanation: \_\_\_Answer: \_\_\_\_''<|im\_end|>\\
    <|im\_start|>\mybox[yellow]{user}\\
    Fact: Cookie files are simple text files that can be viewed in Windows Notepad.\\
    Question: What type of application can be used to open and view cookie files on a Windows computer?\\
    Options:\\A) Microsoft Excel B) Adobe Photoshop C) Windows Notepad D) 3D Modeling Software<|im\_end|>\\
    <|im\_start|>\mybox[red]{assistant}\\
    Explanation: Windows Notepad is a text editor that is capable of opening and displaying the contents of simple text files, which is the format of cookie files.\\
    Answer: C) Windows Notepad<|im\_end|>
\end{tcolorbox}

\subsection{R-example}
\begin{tcolorbox}[fontupper = \ttfamily]
    <|im\_start|>\mybox[green]{system}\\
    You are a cautious assistant. You carefully follow instructions. You are given a question together with a few options, you should give an explanation first and then answer the question. Your response should follow the format like ``Explanation: \_\_\_Answer: \_\_\_\_''<|im\_end|>\\
    <|im\_start|>\mybox[yellow]{user}\\
    Question: What type of application can be used to open and view cookie files on a Windows computer?\\Options:\\A) Microsoft Excel B) Adobe Photoshop C) Windows Notepad D) 3D Modeling Software<|im\_end|>\\
    <|im\_start|>\mybox[red]{assistant}\\
    Explanation: Windows Notepad is a text editor that is capable of opening and displaying the contents of simple text files, which is the format of cookie files.\\
    Answer: C) Windows Notepad<|im\_end|>
\end{tcolorbox}

\section{Templates for evaluation}
\label{app:evaluation}
\subsection{LLaMA-2, LLaMA-3, and MeanLearn (8B)}
\begin{tcolorbox}[fontupper = \ttfamily]
    The following are multiple-choice questions (with answers) about abstract algebra.\\\\
    Find the degree for the given field extension Q(sqrt(2), sqrt(3), sqrt(18)) over Q.\\
    A. 0 B. 4 C. 2 D. 6\\
    Answer: A
\end{tcolorbox}

\subsection{Orca-2 and MeanLearn (7B and 13B)}
\begin{tcolorbox}[fontupper = \ttfamily]
    <|im\_start|>\mybox[green]{system}\\
    The following are multiple-choice questions (with answers) about abstract algebra.<|im\_end|>\\
    <|im\_start|>\mybox[yellow]{user}\\
    Question: Find the degree for the given field extension Q(sqrt(2), sqrt(3), sqrt(18)) over Q.\\
    Options: \\
    A. 0 B. 4 C. 2 D. 6<|im\_end|>\\
    <|im\_start|>\mybox[red]{assistant}\\
    Answer: A<|im\_end|>
\end{tcolorbox}

\section{Training details of RoBERTa-Large-based clusterer}
\label{app:roberta}
We use the RoBERTa-Large~\cite{liu2019roberta} released by Meta. The batch size is 256. We use AdamW~\cite{kingma2014adam} to optimize the model with a learning rate of 1e-5 for 5 epochs. The computing device is one NVIDIA A100-SXM-64GB. The running time is about 12 minutes.

\end{document}